\documentclass[runningheads]{llncs}

\usepackage[mobile]{eccv}

\usepackage{eccvabbrv}

\usepackage{graphicx}
\usepackage{booktabs}

\usepackage[accsupp]{axessibility}  % Improves PDF readability for those with disabilities.

\usepackage[pagebackref,breaklinks,colorlinks,citecolor=eccvblue]{hyperref}
\usepackage{orcidlink}

\usepackage{enumitem}
\usepackage{multirow}
\usepackage{colortbl}
\usepackage{arydshln}       % 负责画虚线的包

\usepackage{float}
\usepackage{amsmath}
\usepackage{makecell}
\definecolor{mygray}{gray}{0.6}
\newcommand{\pub}[1]{\color{mygray}{\scriptsize{[{#1}]}}}
\usepackage{wrapfig}

\usepackage{appendix}

\begin{document}

\title{Collaborative Vision-Text Representation Optimizing for Open-Vocabulary Segmentation} 

\titlerunning{MAFT+}

% \author{Siyu Jiao\inst{1}\orcidlink{0000-0002-0795-8401} \and
% Second Author\inst{2,3}\orcidlink{1111-2222-3333-4444} \and
% Third Author\inst{3}\orcidlink{2222--3333-4444-5555}}
% 引连续学， 共一，pengcheng  ^{$\star$}
\author{%
  Siyu Jiao\inst{1,2}\thanks{Equal contribution}\orcidlink{0000-0002-0795-8401} \and 
  Hongguang Zhu\textsuperscript{1,2$\star$}\orcidlink{0000-0002-1356-5153} \and Jiannan Huang\inst{1,3}\orcidlink{0009-0002-2447-9928} 
  \and Yao Zhao\inst{1,2}\orcidlink{0000-0002-8581-9554} 
  \and \\
   Yunchao Wei\inst{1,2}\orcidlink{0000-0002-2812-8781} 
   \and Humphrey Shi \inst{3,4}\orcidlink{0000-0002-2922-5663}
}

\authorrunning{J.~Author et al.}

\institute{Institute of Information Science, Beijing Jiaotong University \and
Peng Cheng Laboratory \and Georgia Institute of Technology \and Picsart AI Research (PAIR) \\
\email{jiaosiyu99@bjtu.edu.cn}}

\maketitle

\begin{abstract}
Pre-trained vision-language models, \textit{e.g.} CLIP, have been increasingly used to address the challenging Open-Vocabulary Segmentation (OVS) task, benefiting from their well-aligned vision-text embedding space. Typical solutions involve either freezing CLIP during training to unilaterally maintain its zero-shot capability, or fine-tuning CLIP vision encoder to achieve perceptual sensitivity to local regions.  However, few of them incorporate vision-text collaborative optimization. 
Based on this, we propose the Content-Dependent Transfer to adaptively enhance each text embedding by interacting with the input image, which presents a parameter-efficient way to optimize the text representation. Besides, we additionally introduce a Representation Compensation strategy, reviewing the original CLIP-V representation as compensation to maintain the zero-shot capability of CLIP.
In this way, the vision and text representation of CLIP are optimized collaboratively, enhancing the alignment of the vision-text feature space. To the best of our knowledge, we are the first to establish the collaborative vision-text optimizing mechanism within the OVS field. 
Extensive experiments demonstrate our method achieves superior performance on popular OVS benchmarks. In open-vocabulary semantic segmentation, our method outperforms the previous state-of-the-art approaches by +0.5, +2.3, +3.4, +0.4 and +1.1 mIoU, respectively on A-847, A-150, PC-459, PC-59 and PAS-20. Furthermore, in a panoptic setting on ADE20K, we achieve the performance of 27.1 PQ, 73.5 SQ, and 32.9 RQ.
Code will be available at 
\href{https://github.com/jiaosiyu1999/MAFT-Plus.git}{MAFT-Plus}.
\keywords{Open-Vocabulary Segmentation \and Fine-tuning}
\end{abstract}

\section{Introduction}
\label{sec:intro}
Segmentation stands as the most popular basic topics in computer vision, traditional segmentation models \cite{chen2017deeplab,pspnet,huang2019ccnet,huang2021alignseg, fang2023locating} are only capable of segmenting a few predefined categories within a closed vocabulary \cite{pascal, coco}, notably smaller than the human-used categories for describing the real world \cite{zhang2023controlvideo}. Therefore, open-vocabulary segmentation (OVS) \cite{spnet, zs5, cagnet, han2023global} is introduced to segment objects using arbitrary categories described by texts.

\begin{figure}[t]
% \vspace{-10mm}
\begin{center}
   \includegraphics[width=0.99\linewidth]{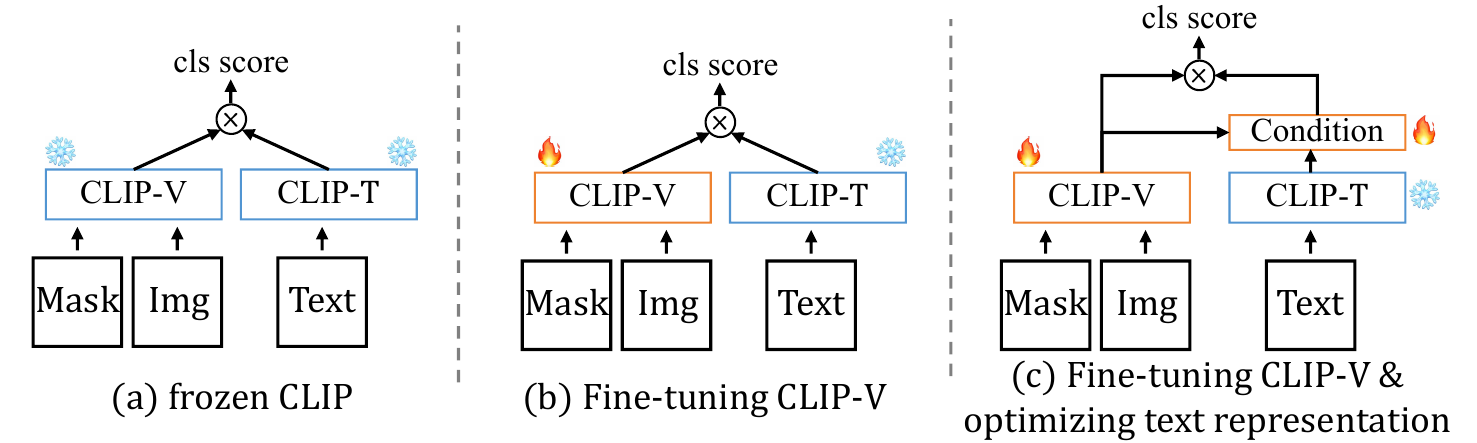}
\end{center}
% \vspace{-3mm}
   \caption{
     Different learning frameworks for open-vocabulary segmentation, from the perspective of whether to freeze CLIP. (a) The "frozen CLIP" paradigm. \cite{ovseg, zsseg, freeseg, fcclip} (b) Fine-tuning CLIP-V \cite{maft}. (c) Our MAFT+ framework enables to optimize both CLIP-V and CLIP-T.
   }
\label{fig:intro}
% \vspace{-5mm}
\end{figure}

Recently, large-scale visual-language pre-training models (\textit{e.g.} CLIP \cite{radford2021learning} and ALIGN \cite{jia2021scaling}) learn representation with cross-modal alignment and show strong zero-shot capability, leading to the increased adoption for tackling the challenging OVS task \cite{zegformer,zsseg,ovseg,freeseg}. A mainstream solution follows the "decoupling" paradigm, which executes the open-vocabulary segmentation with two steps: 1) employing a Proposal Generator to produce class-agnostic mask proposals and 2) leveraging a pre-trained CLIP to classify each mask proposal via similarity matching in the aligned image-text feature space. 
The above-mentioned paradigm can be categorized into two groups hinges on whether CLIP is frozen during the training process, as depicted in Fig. \ref{fig:intro}\textcolor{red}{a}, \textcolor{red}{b}.
% While acceptable results are obtained, we reveal that these approaches overlook a crucial issue, \textit{i.e.} the frozen CLIP is insensitive to different mask proposals and tends to produce similar predictions for various proposals of the same image. 

In order to retain the strong zero-shot capability of CLIP when classifying mask proposals, most previous works \cite{zsseg, ovseg, freeseg, fcclip} choose to freeze the pre-trained CLIP model (Fig. \ref{fig:intro}\textcolor{red}{a}). They execute with either masked-crops or masked-attention, when processing images and masks within CLIP-V. Considering the domain gap between image-level pre-training of CLIP and pixel-level application of segmentation, these approaches compromise the representational ability of CLIP, and fail to fit the distribution of segmentation tasks well.
Recent work MAFT \cite{maft} highlights the frozen CLIP is insensitive to different mask proposals and often yields similar predictions. It designs a mask-aware fine-tuning strategy to enhance the sensitivity of CLIP-V to local regions (Fig. \ref{fig:intro}\textcolor{red}{b}). While MAFT partially addresses the insensitivity issue, it comes with some new problems: 
% \textit{i.e.}, the generalization ability of CLIP deteriorates during CLIP fine-tuning:
1) only updating CLIP-V constrains the overall optimization space, thereby limiting the alignment of vision and text representation. 2) fine-tuning CLIP-V on downstream datasets leads to the degradation of generalization ability.
% \textit{i.e.}, the generalization ability of CLIP deteriorates during CLIP fine-tuning: 1) MAFT shifts CLIP-V towards fixed text embeddings, resulting in a limited optimization space. 2) Fine-tuning CLIP-V parameters on downstream datasets leads to the erosion of its pre-trained zero-shot capabilities.
% MAFT lead to clip generalization下降 （核心问题）：1) MAFT将CLIP-V向固定类别的CLIP-T特征空间的对齐，这种单向的优化导致CLIP-V优化空间较小，并且容易overfit到几个固定类别的seen上。2）优化CLIP-V的参数本身就会导致pre-trained的zero-shot能力的遗忘
% 一方面，另一方面
% 1) MAFT shifts CLIP-V towards fixed text embeddings generated by CLIP-T. This \textcolor{blue}{one-way optimization} results in a limited optimization space and is prone to overfitting the training categories. 2) Fine-tuning CLIP-V parameters on downstream datasets leads to the erosion of its pre-trained zero-shot capabilities.

% In our ongoing exploration of CLIP tuning techniques, we discover that incorporating the CLIP-T during tuning offers a more refined solution. (Fig. \ref{fig:intro} (c)) By performing this approach, two advantages are provided: 1) Incorporating the CLIP Text Encoder during tuning optimizes the alignment of CLIP image and text features, reduces the complexity of unilaterally training CLIP. \textcolor{blue}{res} 2) \textcolor{blue}{Incorporating the CLIP Text Encoder during tuning diversifies the CLIP text feature, preventing CLIP Vision Encoder from overfitting to the fixed text-embedding during the mask-aware tuning process.

To address the aforementioned problems, we introduce a collaborative Vision-Text representation fine-tuning framework as the enhanced version of MAFT, named MAFT+. As shown in Fig. \ref{fig:intro}\textcolor{red}{c}.
Specific to enhance the alignment of vision-text representation, we incorporate CLIP-T into the fine-tuning process to concurrently optimize the text representation. This vision-text joint optimization alleviates the training complexity and enhances the vision and text alignment. Considering the challenging GPU memory requirements for fine-tuning CLIP-T, we introduce a Content-Dependent Transfer (CDT) following CLIP-T to optimize text representation in a parameter-efficient way. CDT utilizes Transformer Layers to condition text embeddings on each input image rather than fixed once generated by CLIP-T, mitigating the computational burden while preserving the effectiveness of the fine-tuning process.
Moreover, to maintain the zero-shot capality during CLIP-V fine-tuning, we draw inspiration from preventing Catastrophic Forgetting \cite{mccloskey1989catastrophic} in continual learning, and devise a Representation Compensation (RC) strategy. This strategy aims to preserve CLIP's zero-shot capability by reviewing the pre-trained representation of an original CLIP-V as a form of compensation.
% By performing MAFT+, two advantages are provided: 1) 
% % MAFT focuses solely on training CLIP-V, confining the overall optimization space. Whereas 
% MAFT+ employs joint optimization, allowing for further exploration of the optimization potential. 2) The Representation Compensation strategy effectively maintains the original zero-shot capability throughout the CLIP-V training process.

Overall, our contributions are summarized as follows:
\begin{itemize}[itemsep=2pt,topsep=0pt,parsep=0pt]
\item Our MAFT+ represents the first collaborative framework to jointly optimize vision-text representation in OVS. This collaborative design mitigates training complexity and enhances alignment in the vision-text feature space. %As demonstrated in Fig. \ref{fig:intro}\textcolor{red}{c}.
\item The Content-Dependent Transfer is proposed to unleash the optimization potential of CLIP-T through parameter-efficient fine-tuning. The Representation Compensation achieves effective CLIP-V fine-tuning while maintaining the original zero-shot capability.  
% MAFT+ achieves superior performance on OVS benchmarks within an end-to-end framework.
% 图像自适应感知text space

% \item We introduce a Representation Compensation (RC) strategy to maintain CLIP's zero-shot capability, which reviews the pre-trained zero-shot knowledge of the original CLIP-V as compensation during training. The Content-Dependent Transfer (CDT) is proposed to apply parameter-efficient CLIP-T fine-tuning. The optimization of CLIP-T representation facilates the alignment of vision and text embeddings.
% \item Our proposed method, MAFT+, achieves the superior performance on multiple open-vocabulary segmentation datasets.
\end{itemize}

% We conduct extensive experiments to evaluate MAFT+, 
We evaluate our MAFT+ on the commonly used open-vocabulary \textit{semantic} and \textit{panoptic} segmentation benchmarks: Pacal-Context \cite{pc}, Pascal-VOC \cite{pascal}, and ADE20K \cite{ade20k}. Compared with the prior open-vocabulary \textit{semantic} results, MAFT+ enhances the performance of A-847 \cite{ade20k}, A-150 \cite{ade20k}, PC-459 \cite{pc}, PC-59 \cite{pc} and PAS-20 \cite{pascal} datasets by +0.5, +2.3, +3.4, +0.4 and +1.1 mIoU respectively. Furthermore, we conduct experiments in a \textit{panoptic} setting, where MAFT+ achieves the performance of 27.1 PQ, 73.5 SQ, and 32.9 RQ on the ADE20K dataset.
Notably, our approach outperforms the existing OVS methods and establishes new state-of-the-art results across all evaluated datasets.
%前两段没太大问题，3,4段要大改，可能要重新整理思路，站在论文全局去想，扣住自己的优势和之前方法的痛点。可以先总结贡献点作为提纲去写3,4段。
% 贡献点：

\section{Related Work}
\label{sec:related}
\noindent \textbf{Open-Vocabulary Segmentation}~\cite{shaban2017one} is established to break category restrictions and perform segmentation across arbitrary categories.
Earlier works \cite{spnet, zs5, cagnet, li2022languagedriven, xu2022groupvit} use large pre-trained vision-language models to perform open-vocabulary segmentation, they leverage rich alignment features from image-text pairs. Recent approaches \cite{zegformer, zsseg, ovseg, freeseg, ghiasi2022scaling, san, odise, opsnet, fcclip, maft, xu2024transferable} decouple the open-vocabulary segmentation into mask proposals generation and mask proposals classification, they first generate a series of mask proposals and then utilize CLIP \cite{radford2021learning} or ALIGN \cite{jia2021scaling} for classification. 
Specifically, Zegformer \cite{zegformer} first uses mask\&crop to get sub-images based on mask proposals, feeding them into CLIP for mask classification. The following approaches ZSSeg \cite{zsseg} and OVSeg \cite{ovseg}, train CLIP adapters to boost performance. 
In order to improve the classification ability of the vision-language models, OpenSeg \cite{ghiasi2022scaling} takes extra image-caption pairs to scale up training data. FreeSeg\cite{freeseg} unifies semantic, instance, and panoptic tasks and performs fusion training. ODISE \cite{odise} utilizes a strong text-to-image diffusion model \cite{rombach2022high} to obtain a well-aligned image-text feature space. 
SAN \cite{san} and FC-CLIP \cite{fcclip} design the end-to-end frameworks by exploiting a single frozen CLIP as the backbone. Recently, MAFT \cite{maft} introduces a CLIP-V fine-tuning strategy, allowing CLIP-V to be sensitive to different mask proposals.
% However, the freezing CLIP reduces the representational ability of the model, and does not fit the distribution to zero-shot segmentation task.

% \noindent \textbf{\textcolor{blue}{Pre-trained model fine-tuning}}
% is widely used for fitting the distribution to downstream tasks, \textit{e.g.} detection and segmentation. 
% \cite{zhou2022learning, zhou2022conditional, guo2022texts, zsseg, ovseg, freeseg, lion, tpt, detpro} apply prompt-tuning to learn text prompts or image prompts by annotated data from the target tasks. SVF \cite{svf}  fine-tunes only a few parameters in the pre-trained image encoder to adapt pre-trained knowledge to few-shot segmentation. 
% Some continual learning approaches use contrastive learning \cite{zhangcoinseg, zhang2022mining}, distillation \cite{conti-dis} and EMA \cite{conti-ema} to avoid catastrophic forgetting.
% DetPro \cite{detpro} performs open-vocabulary detection by learning continuous prompt representations.

\noindent \textbf{Pre-trained model fine-tuning}
is widely used for fitting the distribution to downstream tasks. Specific to segmentation, traditional close-set methods \cite{chen2017deeplab,pspnet,huang2019ccnet,huang2021alignseg} typically use a lower learning rate (\textit{e.g.} $\frac{1}{10}$) to fine-tune the image encoder, transferring pre-trained knowledge to segmentation tasks. However, this strategy may be suboptimal for data-limited scenarios such as few-shot segmentation, zero-shot segmentation and incremental segmentation due to the daunting \textit{overfitting} problem.
To tackle this, SVF \cite{svf}  fine-tunes only a subset of parameters in the pre-trained image encoder, adapting pre-trained knowledge to few-shot segmentation. 
\cite{ovseg} applies prompt-tuning to learn image prompts using annotated data, adapting CLIP-V to masked images. 
Some continual segmentation approaches utilize techniques like contrastive learning \cite{zhangcoinseg, zhang2022mining, zhang2023slca}, distillation \cite{conti-dis} and EMA \cite{conti-ema} to avoid catastrophic forgetting.

In a recent development, MAFT \cite{maft} conducts a mask-aware CLIP fine-tuning strategy by aligning CLIP's classification score with the IoU score. 
Although this approach partially adapts CLIP-V to segmentation tasks, it exclusively optimizes CLIP-V representation, potentially amplifying the training difficulty and risking overfitting on fixed text embeddings.
This observation motivates our exploration of collaborative optimization strategies for both vision and text representation.

% DetPro \cite{detpro} performs open-vocabulary detection by learning continuous prompt representations.

\section{Preliminary}
\label{sec:prelimiary}

% \subsection{Problem Setting}
\noindent \textbf{Problem Setting.}
Open-vocabulary segmentation addresses the task of training a segmentation model capable of segmenting arbitrary objects using text descriptions. Given two category sets $C_{train}$ and $C_{test}$, where $C_{train}$ and $C_{test}$ are unequal in terms of object categories ($C_{train} \neq C_{test}$). The model is trained on $C_{train}$ and directly tested on $C_{test}$. Typically, $C_{train}$ and $C_{test}$ are described by noun words (\textit{e.g.} sky, sea, mount...).

\noindent \textbf{\textit{mask-aware} Loss Function.}
\cite{maft} proposes a \textit{mask-aware} loss ($\mathcal{L}_{ma}$) to fine-tune CLIP-V for sensitivity to local regions. The primary objective of $\mathcal{L}_{ma}$ is to assign high classification scores to high-quality proposals and low scores to low-quality proposals. 
This is achieved by utilizing the Intersection over Union (IoU) score $S^{IoU}$ derived from ground-truth as supervision and aligning it with the CLIP classification score $S^{cls}$ to induce mask awareness. The \textit{mask-aware} loss is calculated using the $\mathrm{SmoothL1}$ function:
\begin{equation}
\mathcal{L}_{ma} = \mathrm{SmoothL1} (S^{cls}, S^{IoU})
   \label{con:lma}
\end{equation}
% Although  "frozen CLIP" approaches have achieved promising results, it is clear that directly adopting an image-level pre-trained CLIP for proposal classification can be suboptimal. A frozen CLIP usually produces numerous false positives, and the \textit{merge} operation may destroy the context information of an input image. 
In this paper, we use $\mathcal{L}_{ma}$ to fit the distribution of CLIP with OVS. Furthermore, we delve into CLIP fine-tuning techniques, and propose a novel CLIP fine-tuning strategy by collaboratively optimizing the distribution of CLIP-V and CLIP-T.
% MAFT+ shares a goal similar to that of MAFT. \textit{i.e.}, fine-tuning CLIP to adapted pixel-level tasks and fit CLIP's distribution to open-vocabulary segmentation. 
% In view of this, we propose a novel CLIP fine-tuning strategy by collaboratively optimizing the feature distribution of CLIP-V and CLIP-T.

\section{Methodology}
\label{sec:method}
\begin{figure}[t]
% \vspace{5mm}
\begin{center}
   \includegraphics[width=0.99\linewidth]{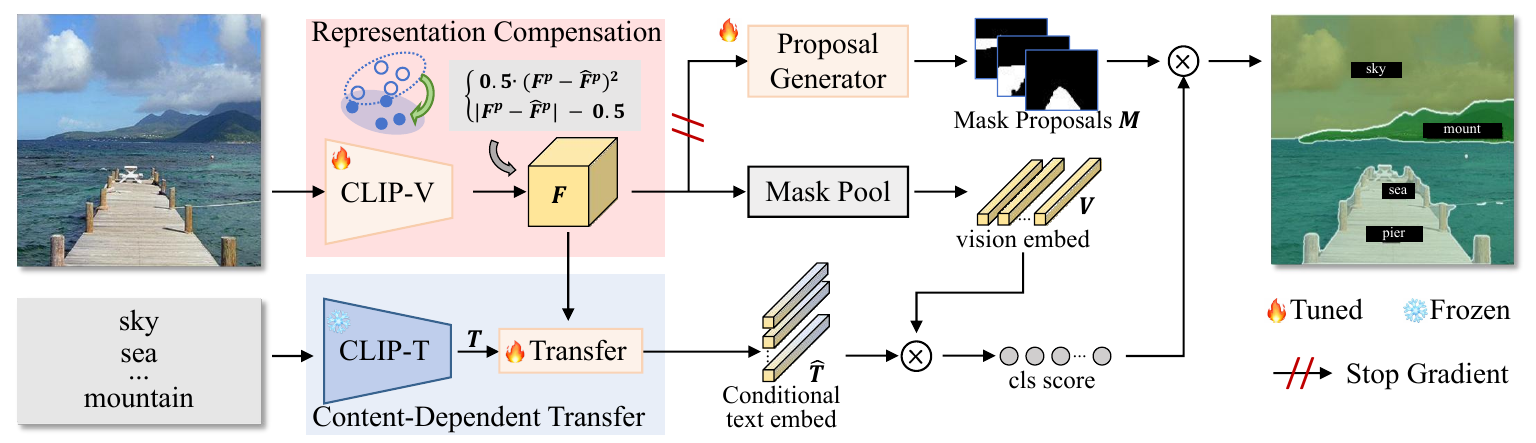}
\end{center}
% \vspace{-3mm}
   \caption{
    Overview of the MAFT+. We use CLIP-V as the backbone to extract image features. A Proposal Generator is trained to generate mask proposals. The Representation Compensation strategy reviews the vision representation to preserve the zero-shot capability of CLIP (\textcolor{red}{red} part); the Content-Dependent Transfer enables the text embeddings conditioned on input image, and achieves text representation optimizing in a parameter-efficient fine-tuning way. (\textcolor{cyan}{blue} part).
   }
\label{fig:framework}
% \vspace{-5mm}
\end{figure}

We introduce MAFT+, a method for collaboratively optimizing CLIP's vision and text representation. The complete framework of the MAFT+ is shown in Fig.~\ref{fig:framework}, we use the Convnext-Large CLIP model for illustration.
Within MAFT+, CLIP-V serves as the vision backbone, and a Proposal Generator is trained to generate class-agnostic mask proposals (Sec. \ref{sec:PG}). 
Simultaneously, the representation of CLIP-V and CLIP-T is collaboratively optimized. We introduce the Representation Compensation (RC) strategy for CLIP-V fine-tuning (Sec. \ref{sec:RC}), and propose the Content-Dependent Transfer (CDT) for parameter-efficient CLIP-T fine-tuning (Sec. \ref{sec:CDT}). 
Finally, we outline the loss functions in Sec. \ref{sec:loss}.

\subsection{Feature Extraction \& Proposal Generator}
\label{sec:PG}

\noindent \textbf{Feature Extraction.}
We utilize a pre-trained convolutional CLIP-V for extracting features from an input image $I$. Denoting each stage of CLIP-V's output as $F=\{F^i\}, i \in [0,1,2,3]$. $F^0$, $F^1$, $F^2$, $F^3$ have strides of \{4, 8, 16, 32\} with respect to the input image. 

\noindent \textbf{Proposal Generator.}
We follow the common design \cite{zegformer, zsseg, ovseg, freeseg, mmformer, fcclip, maft} to use MaskFormer \cite{cheng2021maskformer, cheng2021mask2former} as the Proposal Generator. Since the Hungarian matching \cite{kuhn1955hungarian} is used in the training process, only a subset of the mask proposals is optimized. This matching strategy enhances generalizability of the Proposal Generator, ensuring it segment masks of novel categories. Given the image features $F$, the Proposal Generator generates a set of $N$ mask proposals $M=\{m_i\}^N_{i=1} \in \mathbb{R}^{N \times H \times W}$.

During the training process, we stop the gradient flow from CLIP-V to the Proposal Generator. This measure is taken to avoid the potential overfitting of CLIP-V on the training categories.

\subsection{Representation Compensation}
\label{sec:RC}

\begin{wrapfigure}{r}{0.45\textwidth}
% \vspace{-5mm}
\vspace{-12mm}
\begin{center}
\includegraphics[width=0.45\textwidth]{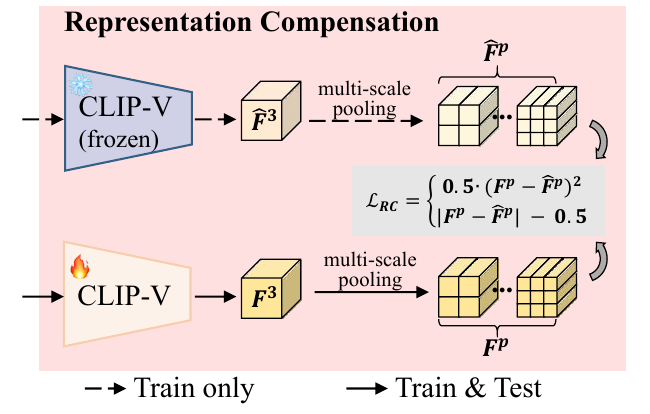}
\end{center}
\vspace{-5mm}
\caption{Details of Representation Compensation.}
\label{fig:RC}
\vspace{-5mm}
\end{wrapfigure}

The representation Compensation (RC) strategy aims to review the original representation of CLIP as compensation during the training phase. 
Details of Representation Compensation are shown in Fig. \ref{fig:RC}. Within RC, 
% we achieve representation compensation by constraining the features of the fine-tuned CLIP-V to be consistent with the features of the original CLIP-V. Specifically, 
we use a frozen CLIP-V (denoted as CLIP-V*) to generate the original CLIP-V features during training. Extracting the last stage output from the CLIP-V* ($\hat{F}^3$) and the fine-tuned CLIP-V ($F^3$), $\hat{F}^3$ and $F^3$ are expected to be similar to avoid Catastrophic Forgetting. However, direct per-pixel alignment is not feasible, as it would result in the loss of region-level differences. Therefore, we devise multiple grids of average pooling ($\mathrm{AvgPooling}$) to generate multi-scale features, and ensure the consistency of the features after pooling. 

Given an arbitrary feature $f \in \mathbb{R}^{d \times h \times w}$, an $\mathrm{AvgPooling}$ operation with grid size of $k \times k$ can be formulated as:
\begin{equation}
f^{pool} = \mathrm{AvgPooling} (f, k), f^{pool} \in \mathbb{R}^{d \times k \times k}.
\end{equation}
In our default design, we use $\mathrm{AvgPooling}$ with $K=\{1,2,4\}$ to perform pooling $\hat{F}^3$ and ${F}^3$ into $\{1\times 1, 2\times 2, 4\times 4\}$ grids, denoting as $\hat{F}^p$ and ${F}^p$. Specifically, $\hat{F}^p = \mathrm{AvgPooling} (\hat{F}^3, K)$ and ${F}^p = \mathrm{AvgPooling} ({F}^3, K)$. Then, we use $\mathrm{SmoothL1}$ Loss to minimize the difference as follows:
\begin{equation}
\mathcal{L}_{rc} = \mathrm{SmoothL1} ({F}^p, \hat{F}^p) ,
\end{equation}
\begin{equation}
\mathrm{SmoothL1}({F}^p, \hat{F}^p) = \left\{
\begin{aligned}
 0.5\cdot ({F}^p - \hat{F}^p)^2  &, ~~~ \mathrm{if} ~ |{F}^p - \hat{F}^p| < 1\\
|{F}^p - \hat{F}^p| - 0.5  &, ~~~ \mathrm{otherwise} ~ \\ 
\end{aligned}
\right.
\end{equation}
With RC to compensate ${F}^3$ original CLIP's representation, the CLIP-V maintains the zero-shot capability during fine-tuning. We apply $\mathrm{Mask}$ $\mathrm{Pooling}$ \cite{fcclip} on the ${F}^3$ to generate vision embeddings ($V \in \mathbb{R}^{N \times d}$) for each mask proposal.

\subsection{Content-Dependent Transfer}
\label{sec:CDT}
Given a set of class names $C=\{C_1, C_2...C_n\}$, we use the predefined templates \cite{zsseg, san, odise, fcclip, maft} to generate sentences corresponding to these class names, \textit{e.g.}, "a photo of a $\{C_i\}$; There is a $\{C_i\}$ in the scene...", these sentences are then fed into CLIP-T to generate embeddings of each sentence. The embeddings of the same classes are averaged to obtain text embedding ($T \in \mathbb{R}^{d \times |C|} $). $d$ is the dimension of the embedding, and $|C|$ is the number of class names.

\begin{wrapfigure}{r}{0.45\textwidth}
\vspace{-5mm}
\begin{center}
\includegraphics[width=0.45\textwidth]{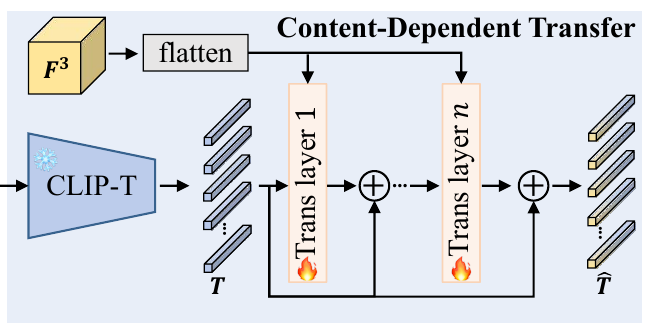}
\end{center}
\vspace{-5mm}
\caption{Details of Content-Dependent Transfer.}
\label{fig:CD}
\vspace{-5mm}
\end{wrapfigure}

To optimize CLIP-T representation $T$, we propose the Content-Dependent Transfer (CDT), which involves a sequence of Transformer Layers performing cross-attention with vision feature $F^3$. Details of the CDT are illustrated in Fig. \ref{fig:CD}.
We take the last stage feature of CLIP-V ($F^3$) and the text embeddings $T$ as the inputs for CDT. $F^3$ is first $\mathrm{Flatten}$ at spatial dimension, denoted as $F^3_{flat} \in \mathbb{R}^{d \times hw}$. Then, we use $n$ sequential Transformer Layers to process $T$ and $F^3_{flat}$, while incorporating a shortcut connection. This process can be formulated as:
\begin{equation}
   T_{i+1} =\mathrm{TransLayer_i}(T_i, F^3_{flat}) + T_i, ~~~i={1,2...l} .
   \label{con:cd}
\end{equation}
In our default setting, $l$ is set to 2. The resulting output of the CDT is denoted as the conditioned text embeddings ($\hat{T}$). Specifically, 
\begin{equation}
   \mathrm{TransLayer}(a, b) =\mathrm{Softmax}(\frac{\mathrm{Que}(a) \cdot \mathrm{Key}(b)}{\sqrt{d}}) \cdot \mathrm{Val}(b) ,
   \label{con:tlayers}
\end{equation}
where $\mathrm{Que}(\cdot)$, $\mathrm{Key}(\cdot)$, and $\mathrm{Val}(\cdot)$ represent linear projections, $d$ is the dimension of the input vectors, we assume all vectors have the same dimension $d$ by default. In Eq. \ref{con:tlayers}, we simplify the expression of Multihead Attention and LayerNorm in Transformer.
Note that the CLIP-T remains frozen during training, and only the Transformer Layers are trained to optimize the CLIP-T representation. Therefore, the parameter-efficient CLIP-T fine-tuning is established, with $\hat{T}$ is conditioned on the input images.

We investigate various designs to optimize the CLIP-T representation ($T$), including fine-tuning CLIP-T, training an additional MLP, incorporating description guidance, etc. Further details are presented in Sec. \ref{sec:abla}.

\subsection{Objective}
\label{sec:loss}

After getting the conditional text embeddings $\hat{T}$, we perform matrix multiplication on $\hat{T}$ and $V$ to derive the classification score $S^{cls}$ for the mask proposals. Subsequently, we multiply $S^{cls}$ with $M$ to obtain the final output.

% L_p = f_mask_former（O,y,x,z）

We use the \textit{mask-aware} loss \cite{maft} ($\mathcal{L}_{ma}$, Eq. \ref{con:lma}) on $S^{cls}$ to optimize the representation of both CLIP-V and CLIP-T. Considering the $\mathcal{L}_{ma}$ may induce overfitting on the training categories and reduce the transferability of CLIP, we introduce $\mathcal{L}_{rc}$ (Sec. \ref{sec:RC}) to compensate CLIP's representation during training. Meanwhile, we follow Mask2Former \cite{cheng2021mask2former} to adopt the same loss functions ($\mathcal{L}_{P}$) to train the Proposal Generator without any special design. 
Therefore, the final loss function ($\mathcal{L}$) can be formulated as: $\mathcal{L} = \mathcal{L}_{P} + {\lambda}_1 {\mathcal{L}_{ma}} + {\lambda}_2 {\mathcal{L}_{rc}}$, where $\lambda_1=1$ and $\lambda_2=0.1$.
% \begin{equation}
% \mathcal{L} = \mathcal{L}_{P} + {\lambda}_1 {\mathcal{L}_{ma}} + {\lambda}_2 {\mathcal{L}_{rc}},
%    \label{loss}
% \end{equation}
% where we set the constant $\lambda_1$ and $\lambda_2$ to 1 and 0.1 in our experiments. 

Note that we stop the gradient from CLIP-V to Proposal Generator. The CLIP-V is not optimized by $\mathcal{L}_{P}$.

\noindent \textbf{Modifications in the panoptic setting.}
The $\mathcal{L}_{ma}$ is tailored for semantic segmentation and lacks the ability to capture instance-level information. We explore adapting the $\mathcal{L}_{ma}$
to panoptic segmentation with the following modification. Specifically, when a mask contains multiple instances, we use binary ground-truth (GT) to mask out redundant instances, retaining only the instance with the highest IoU score with GT. This change allows CLIP-V to learn instance-level knowledge, making $\mathcal{L}_{ma}$ applicable to panoptic segmentation.

\section{Experiments}
\label{sec:exp}
\subsection{Setting}
\noindent \textbf{Dataset.}
We conduct experiments on popular open-vocabulary segmentation benchmarks, including COCO-Stuff, COCO-Panoptic, Pascal-VOC, Pascal-Context and ADE20K. 
We train MAFT+ on COCO-Stuff and testing on ADE20K (A-847, A-150), Pascal-Context (PC-459, PC-59), and Pascal-VOC (PAS-20) to evaluate the performance of open-vocabulary \textit{semantic} segmentation. 
Then, we evaluate MAFT+ in open-vocabulary \textit{panoptic} settings \cite{odise, opsnet, fcclip}, \textit{i.e.}, training on COCO-Panoptic and testing on ADE20K. \\
More details of the dataset settings are provided in the \textit{Appendix}.

\noindent \textbf{Evaluation Metrics.}
To quantitatively evaluate the performance, we follow standard practice \cite{zegformer, zsseg, ovseg, san, odise, fcclip}. Semantic segmentation results are evaluated with mean Intersection over Union (mIoU) \cite{pascal}. 
Panoptic segmentation results are evaluated with the panoptic quality (PQ), segmentation quality (SQ) and recognition quality (RQ) \cite{kirillov2019panoptic}. 

\noindent \textbf{Implementation details.}
We employ ConvNeXt-Large CLIP from OpenCLIP \cite{ilharco2openclip}. The Proposal Generator is built following the default settings of Mask2Former \cite{cheng2021mask2former}. We set the number of class-agnostic mask proposals to 100 ($N=100$). 
During training, the model is optimized with AdamW optimizer with a weight-decay of 0.05. The learning rate is set to 1 $\times$ 10$^{-5}$ for CLIP-V and 1 $\times$ 10$^{-4}$ for other modules. We use a crop size of 1024 $\times$ 1024. The model is trained for 60,000 iterations on COCO with 4 NVIDIA A100 GPUs.

\subsection{Comparisons with State-of-the-art Methods}

\begin{table}[t]
  \centering
  \footnotesize
  \caption{Open-vocabulary \textit{semantic} segmentation performance. mIoU is used to evaluate the performance. * denotes additional ensemble operation \cite{fcclip} used during testing. }
  % \resizebox{1.0\textwidth}{!}{
    \renewcommand\arraystretch{1.0} % 1.95
    \begin{tabular}{l|c|ccccc}
      % \toprule
      \Xhline{0.7pt}

      & VLM & \textbf{A-847} & \textbf{A-150} & \textbf{PC-459} & \textbf{PC-59} & \textbf{PAS-20}\\ 
      \hline
      OpenSeg \pub{ECCV22}\cite{ghiasi2022scaling} & ALIGN & 8.8 &28.6&12.2& 48.2 & 72.2\\ 
      OVSeg \pub{CVPR23}\cite{ovseg} & ViT-L & 9.0 & 29.6 & 12.4 & 55.7 & 94.5\\
      SAN \pub{CVPR23}\cite{san} & ViT-L & 12.4 & 32.1 &  15.7  &  57.7 & 94.6 \\     
      ODISE \pub{CVPR23}\cite{odise} & ViT-L & 11.1 & 29.9 &  14.5  &  57.3 &  - \\     
      FC-CLIP \pub{NeurIPS23}\cite{fcclip} &~~ConvNeXt-L~~&  11.2 & 26.6 &  12.7  &  42.4 & 89.5 \\   
      FC-CLIP* \pub{NeurIPS23}\cite{fcclip} &ConvNeXt-L&  14.8 & 34.0 &  18.2  &  58.4 & 95.4 \\   
      MAFT \pub{NeurIPS23}\cite{maft} & ViT-L &  12.7 & 33.0 &  16.2  &  59.0 & 92.1 \\  
      MAFT \pub{NeurIPS23}\cite{maft} & ConvNeXt-L &  13.1 & 34.4 &  17.0  &  57.5 & 93.0 \\  
      % \hline
      \rowcolor{gray!10}\multicolumn{1}{l|}{MAFT+ (ours)} &ConvNeXt-L& \textcolor{red}{15.1} & \textcolor{red}{36.1} & \textcolor{red}{21.6} & \textcolor{red}{59.4} & \textcolor{red}{96.5} \\
      % \rowcolor{gray!10}\multicolumn{1}{l|}{MAFT+ (ours)} &ConvNeXt-L & 15.3 & 36.3 & 21.6 & 59.4 & 96.5 \\
      \Xhline{0.7pt}
      % \bottomrule
      \end{tabular}
      % }
      % \end{threeparttable}
      \label{tab:ovs}
  % \vspace{-2mm}
  \end{table}

\begin{table}[t]
  \centering
  \footnotesize
  \vspace{-2mm}
  \caption{Open-vocabulary \textit{panoptic} segmentation performance on ADE20K. PQ, SQ, and RQ are used for evaluation. The best results are highlighted with \textcolor{red}{red}. }
  % \resizebox{1.0\textwidth}{!}{
    \renewcommand\arraystretch{1.0} % 1.95
    \begin{tabular}{l|ccc}
      \Xhline{0.7pt}
    \hline
      % & PQ & SQ & RQ \\ 
      & ~~~~PQ~~~~ & ~~~~SQ~~~~ & ~~~~RQ~~~~ \\ 
      \hline
      FreeSeg \pub{CVPR22}\cite{freeseg} & 16.3 & - & - \\
      ODISE \pub{CVPR22}\cite{odise} & 22.6 & - & - \\
      MaskCLIP \pub{ICML23}\cite{ding2023maskclip} ~~ & 15.1 & 70.4 & 19.2 \\
      OPSNet \pub{ICCV23}\cite{opsnet} & 19.0 & 52.4 & 23.0 \\
      FC-CLIP \pub{NeurIPS23}\cite{fcclip} & 21.9 & 71.5 & 26.4 \\
      FC-CLIP* \pub{NeurIPS23}\cite{fcclip} & 26.8 & 71.5 & 32.2 \\
      % \hline
      \rowcolor{gray!10}\multicolumn{1}{l|}{MAFT+ (ours)}& \textcolor{red}{27.1} & \textcolor{red}{73.5} & \textcolor{red}{32.9} \\
      \Xhline{0.7pt}
      % \bottomrule
      \end{tabular}
      % }
      % \end{threeparttable}
      \label{tab:ovs-pan}
  \end{table}

% \begin{table}[t]
%   \begin{minipage}{1.0\textwidth}
%     \centering
%     \includegraphics[width=\linewidth]{example-image-a}
%     \caption{图的标题}
%     \label{fig:myfigure}
%   \end{minipage}\\
%   \begin{minipage}{1.0\textwidth}
%     \centering
%     \begin{tabular}{|c|c|}
%       \hline
%       列1 & 列2 \\
%       \hline
%       数据1 & 数据2 \\
%       数据3 & 数据4 \\
%       \hline
%     \end{tabular}
%     \caption{表格的标题}
%     \label{tab:mytable}
%   \end{minipage}
% \end{table}
In this section, we compare our proposed MAFT+ with the state-of-the-art open-vocabulary \textit{semantic} segmentation methods and open-vocabulary \textit{panoptic} segmentation methods. 

\noindent \textbf{Comparisons in the \textit{semantic} setting.}
In Tab. \ref{tab:ovs}, we present the performance of MAFT+ on various benchmarks.
MAFT+ demonstrates a significant improvement over existing open-vocabulary segmentation models, achieving a performance boost of +0.5, +2.3, +3.4, +0.4, +1.1 mIoU across A-847, PC-459, A-150, PC-59, and PAS-20, respectively.
Moreover, compared to MAFT \cite{maft}, our MAFT+ eliminates the need for an additional fine-tuned CLIP-V. MAFT+ applies an end-to-end pipeline, facilitating both the training and testing processes.

\noindent \textbf{Comparisons in the \textit{panoptic} setting.}
% The original MAFT \cite{maft} is tailored for semantic segmentation, and $\mathcal{L}_{ma}$ lacks the ability to capture instance-level information. We explore adapting the $\mathcal{L}_{ma}$
% % mask-aware fine-tuning strategy 
% to panoptic segmentation with some simple modifications. Specifically, when a mask contains multiple instances, we use binary ground-truth (GT) to mask out redundant instances, retaining only the instance with the highest IoU score with GT. 
% % \textcolor{blue}{This process only affects the fine-tuning of CLIP-V and does not affect the training of the proposal generator and Content-Dependent Transfer.}
In Tab. \ref{tab:ovs-pan}, we evaluate our MAFT+ on ADE20K, the main evaluation dataset of open-vocabulary panoptic segmentation.  With the aforementioned modifications, our approach achieves new state-of-the-art performance. Compared to FC-CLIP without the ensemble strategy (3$^{rd}$ last results), our MAFT+ outperforms it by +5.2 PQ, +2.0 SQ and +6.5 RQ. Although the ensemble strategy greatly improves FC-CLIP's performance, our model still outperforms FC-CLIP* across all evaluation metrics. 

\noindent \textbf{Analysis of the ensemble strategy in FC-CLIP.}
FC-CLIP ensembles the classification score of Mask2Former and CLIP,  along with two hyper-parameters to balance these scores. As shown in Tab. \ref{tab:ovs} and Tab. \ref{tab:ovs-pan}, the ensemble operation significantly improves FC-CLIP's performance. \textit{i.e.}, 42.4$\rightarrow$58.4 mIoU on PC-59, 21.9$\rightarrow$26.8 PQ on ADE20K. However, \textbf{this improvement stems from the overlap of categories between training and testing datasets}. Moreover, determining the two critical hyper-parameters requires numerous repeated experiments. Based on this, in our default settings, we remove this ensemble operation, and solely use the CLIP for classification.

% To better showcase the performance gains from MAFT, we removed the \textit{ensemble} operation in \cite{zegformer, zsseg, freeseg} and presented the results in Tab. \ref{tab:zss-woensem}.  It can be seen that the performance of different methods is significantly improved after applying MAFT. In particular, the state-of-the-art method FreeSeg achieves hIoU improvements of 19.1\%, 7.0\%, and 8.3\% on COCO, VOC2012 and ADE20K datasets. 

% \noindent \textbf{Comparisons in the \textit{open-vocabulary} setting.}
% We further evaluated the transferability of MAFT in the \textit{open-vocabulary} setting \cite{ovseg, zsseg}, using FreeSeg as a baseline for comparison. Results are shown in Tab. \ref{tab:ovs}.
% Compared with OVSeg \cite{ovseg} and OpenSeg \cite{ghiasi2022scaling}, FreeSeg achieves suboptimal performance. However, the proposed MAFT enhances the performance of A-847, A-150, PC-459, PC-59 and PAS-20 by 3.0\%,11.2\%, 6.4\%, 19.1\% and 4.4\%, and outperforms OpenSeg on all five datasets.

\subsection{Ablation Study}
\label{sec:abla}

We conduct ablation studies on various choices of designs of our MAFT+, and showcase their contribution to the final results in Tab. \ref{tab:component}, \ref{tab:cd}, \ref{tab:RC}. We freeze the CLIP-V and remove the Content-Dependent Transfer as the baseline model (\textit{i.e.} representation of a frozen CLIP). 
% In Tab. \ref{tab:component}, we conduct ablations on Representation Compensation and Content-Dependent Transfer,

\noindent \textbf{Component-wise ablations.} To understand the effect of each component in the MAFT+, including the Representation Compensation (RC) strategy and the Content-Dependent Transfer (CDT). 
We start with a frozen CLIP as the baseline model, and gradually add each design. (Tab. \ref{tab:component}). 
The frozen CLIP yields inferior performance due to CLIP's region-unaware property ($1^{st}$ row). Then, Content-Dependent Transfer optimizes CLIP Text representation and promotes the alignment of vision and text embeddings, resulting in an improvement of +5.8 mIoU on A-150 and +12.8 mIoU on PC-59  ($2^{nd}$ row). 
% However, mask-awareness is not accomplished at this point.
Using only Representation Compensation for fine-tuning CLIP-V produces decent performance (the $3^{rd}$ result), 26.6$\rightarrow$34.8 on A-150, 42.4$\rightarrow$57.1 on PC-59 in terms of mIoU. Finally, introducing CDT and RC collaboratively learns effective vision and text alignment representation, fitting the distribution of CLIP from image-level to segmentation tasks, further enhancing the performance to establish state-of-the-art benchmarks.  (last row).

\begin{table}[t]
  \centering
  \footnotesize
  % \vspace{-10pt}
  \caption{Ablation on components of \textbf{MAFT}+. Here RC and CDT denote Representation Compensation and Content-Dependent Transfer. Note that ``tune CLIP-T'' represents optimizing the distribution of text-embeds, not directly fine-tuning CLIP-T.}
   % \vspace{-10pt}
  % \begin{threeparttable}
  % \resizebox{0.9\textwidth}{
    \renewcommand\arraystretch{1.05} % 1.95
    % \begin{tabular}{l|ccccc}
    \begin{tabular}{l|lllll}
      % \toprule
      \Xhline{0.7pt}

      & \textbf{A-847} & \textbf{A-150} & \textbf{PC-459} & \textbf{PC-59} & \textbf{PAS-20}\\ 
      \hline
      frozen CLIP (baseline)&  ~~11.2 & ~~26.6 &  ~~12.7  &  ~~42.4 & ~~89.5 \\     
      + CDT (tune CLIP-T) &  ~~13.3 $_{\textcolor{red}{+2.1}}$ & ~~32.4 $_{\textcolor{red}{+5.8}}$ &  ~~17.2 $_{\textcolor{red}{+4.5}}$ &  ~~55.2 $_{\textcolor{red}{+12.8}}$ & ~~94.7 $_{\textcolor{red}{+5.2}}$ \\     
      + RC ~~(tune CLIP-V) &  ~~14.6 $_{\textcolor{red}{+3.4}}$ & ~~34.8 $_{\textcolor{red}{+8.2}}$ &  ~~18.2  $_{\textcolor{red}{+5.5}}$ &  ~~57.1 $_{\textcolor{red}{+14.7}}$ & ~~95.3 $_{\textcolor{red}{+5.8}}$ \\     
      % \cdashline{1-6}[0.8pt/2pt]
      % MAFT+ (ours) & ~~\textbf{15.3} & ~~\textbf{36.1} & ~~\textbf{21.6} & ~~\textbf{59.4} & ~~\textbf{96.5} \\
      + CDT \& RC  & ~~15.1 $_{\textcolor{red}{+3.9}}$ & ~~36.1 $_{\textcolor{red}{+9.5}}$ & ~~21.6 $_{\textcolor{red}{+8.9}}$ & ~~59.4 $_{\textcolor{red}{+17.0}}$ & ~~96.5 $_{\textcolor{red}{+7.0}}$ \\
      \Xhline{0.7pt}
      % \bottomrule
      \end{tabular}
      % }
      % \end{threeparttable}
      \label{tab:component}
  \vspace{-2mm}
  \end{table}

\begin{figure}[ht]
\vspace{-6mm}
\begin{center}
   \includegraphics[width=1.0\linewidth]{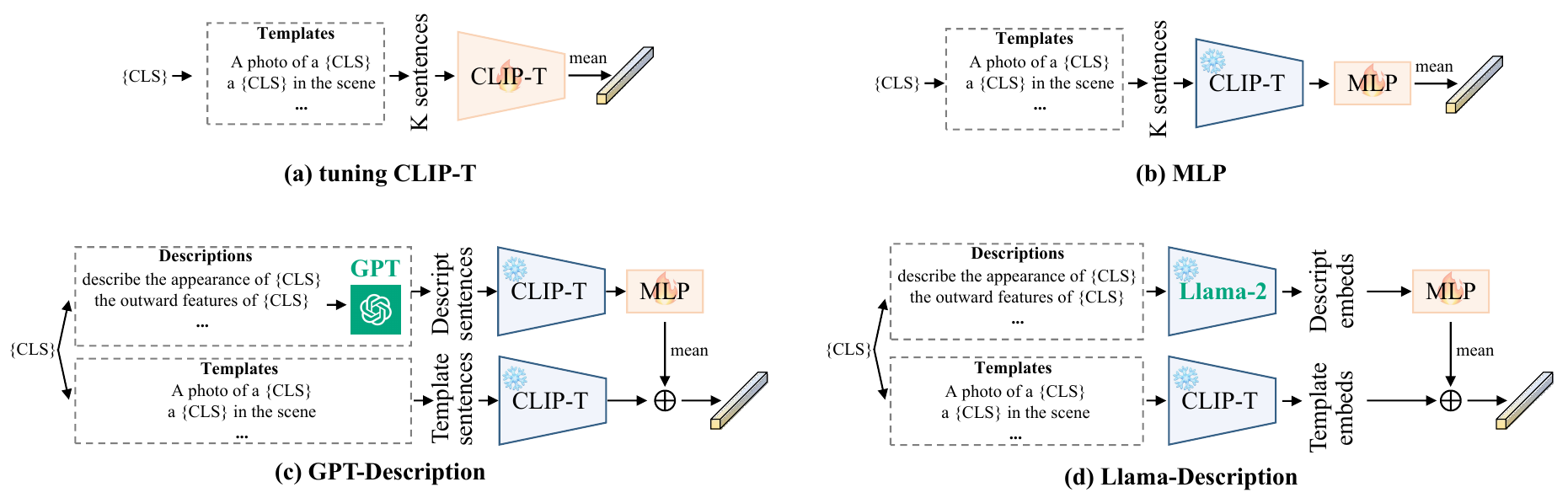}
\end{center}
\vspace{-5mm}
   \caption{
    Comparisons between CLIP-T tuning strategies. 
   }
\label{fig:text}
% \vspace{-5mm}
\end{figure}

\noindent \textbf{Effect of Content-Dependent Transfer.} 
Optimizing CLIP text representation is an essential design of MAFT+. We investigate various  designs to optimize the CLIP-T representation in Fig. \ref{fig:text}, including direct fine-tuning of CLIP-T parameters, training with additional MLP, training with class-description sentences by GPT, and training with class-description embeddings by Llama-2. 
Tab. \ref{tab:cd} presents the results of different designs for optimizing CLIP text representation. Here, we remove Representation Compensation strategy, and keep the CLIP-V frozen for analysis.

\begin{table}[ht]
  \centering
  \footnotesize
  % \vspace{-10pt}
  \caption{Ablation of diverse designs of CLIP-Text optimization. We remove the Representation Compensation strategy and freeze CLIP-V for analysis. Note that fine-tuning CLIP-T requires excessive GPU memory, and thus it is infeasible (denoted as $\mathrm{N/A}$) for the setting in the 2$^{nd}$ row.
}
\vspace{-3mm}
  % \begin{threeparttable}  
  % \resizebox{1.0\textwidth}{
    \renewcommand\arraystretch{1.05} % 1.95
    % \begin{tabular}{l|ccccc}
    \begin{tabular}{l|lllll}
      % \toprule
      \Xhline{0.7pt}

      & \textbf{A-847} & \textbf{A-150} & \textbf{PC-459} & \textbf{PC-59} & \textbf{PAS-20}\\ 
      \hline
      frozen CLIP (baseline)&  ~~11.2 & ~~26.6 &  ~~12.7  &  ~~42.4 & ~~89.5 \\     
      + fine-tune CLIP-T&  ~~$\mathrm{N/A}$ & ~~$\mathrm{N/A}$ &  ~~$\mathrm{N/A}$  &  ~~$\mathrm{N/A}$ & ~~$\mathrm{N/A}$ \\     
      + MLP &  ~~4.1  & ~~20.2 &  ~~11.2  &  ~~51.4 & ~~89.4 \\     
      + GPT-Description &  ~~11.9  & ~~28.2 &  ~~13.3  &  ~~42.6 & ~~90.6 \\     
      + Llama-Description &  ~~9.6  & ~~26.1 &  ~~11.5  &  ~~40.8 & ~~90.9 \\ 
      \rowcolor{gray!10}\multicolumn{1}{c|}{+ Content-Dependent Transfer}
       &  ~~\textcolor{red}{13.3} & ~~\textcolor{red}{32.4} &  ~~\textcolor{red}{17.2} &  ~~\textcolor{red}{55.2} & ~~\textcolor{red}{94.7} \\     
      \Xhline{0.7pt}
      \end{tabular}
      % }
      \label{tab:cd}
  \vspace{-2mm}
  \end{table}

\begin{itemize}[itemsep=2pt,topsep=0pt,parsep=0pt]
\item \textbf{a. fine-tuning CLIP-T}  We explore fine-tuning CLIP-T parameters to optimize the CLIP text representation. The category name ($\{\mathrm{CLS}\}$) is first augmented to sentences by some templates \cite{zsseg, san, odise, fcclip, maft} and fed into CLIP-T However, fine-tuning CLIP-T (2$^{nd}$ results in Tab. \ref{tab:cd}) requires excessive GPU memory (more than 8 NVIDIA A100 GPUs), which is unaffordable in our experiments.
\item \textbf{b. MLP} An MLP layer is added after CLIP-T, with the MLP learning to project text embedding to fit segmentation distributions. Within this design, CLIP-T is frozen, greatly reducing GPU memory consumption compared with fine-tuning CLIP-T. According to the 3$^{rd}$ results in Tab. \ref{tab:cd}, the performance suffers a significant drop on ADE20K (11.2$\rightarrow$4.1, 26.6$\rightarrow$20.2), while increasing on PC-59 (42.4$\rightarrow$51.4). This could be attributed to the MLP layer losing CLIP's zero-shot capability and its inability to perceive novel categories effectively.
\item \textbf{c. GPT-Description} We assume that the detailed description of $\{\mathrm{CLS}\}$ contains additional valuable information, helping to optimize CLIP-T distribution. 
To explore this, we leverage GPT-3.5 \cite{gpt} to generate description sentences of one $\{\mathrm{CLS}\}$. 
\textit{e.g.}, if the instruction provided to GPT is: $[\mathrm{Instruct}]=$``Please describe the appearance of $cat$.'' GPT responds the description sentences of $cat$: $[\mathrm{Response}]=$ ``$[$-a rounded head; -a short snout; -triangular ears ...$]$''
Then we use a frozen CLIP-T to generate the corresponding text embeddings, followed by an MLP layer to project the embeddings. Within this design, the performance is slightly improved: +0.7 on A-847, +2.0 on A-150 (4$^{th}$ results in Tab. \ref{tab:cd}). 
\item \textbf{d. Llama-Description} 
In view of Large Language Models (LLMs) powerful text representation capability, we explore to use of the open-source LLM, Llama-2 \cite{touvron2023llama}, to generate descriptive text embeddings. After obtaining Llama and CLIP-T embeddings, we average them and train an MLP layer to project the Llama embeddings into the CLIP-T embeddings space.  Our experimental results demonstrate that this design does not benefit the performance (5$^{th}$ results in Tab. \ref{tab:cd}). The mIoU drops from 11.2 to 9.6 on A-847, 12.7$\rightarrow$11.5 on PC459. This decrease may be due to the fact that the LLMs' feature space is not aligned with the CLIP-V's visual feature space.
\item \textbf{Content-Dependent Transfer} We propose the Content-Dependent Transfer to enhance CLIP Text embeddings conditioned on the input images. Details can be found in Sec. \ref{sec:CDT}. As shown in the last results in Tab. \ref{tab:cd}, the Content-Dependent Transfer improves the performance on all five datasets: 11.2$\rightarrow$13.3, 26.6$\rightarrow$32.4, 12.7$\rightarrow$17.2, 42.4$\rightarrow$55.2, and 89.5$\rightarrow$94.7, respectively.
\end{itemize}

\textbf{Analysis of why LLMs do not work?} OVS focuses on data-limited settings, examines the model's ability to segment arbitrary text after seeing a few classes. Therefore, effective image-text alignment of prior models (\textit{e.g.}, CLIP) is crucial. Despite LLMs' strong text processing capabilities, their potential is not fully realized with limited data, resulting in incomplete image-text alignment. Thus, simply adapting LLMs to OVS is unsuitable and may require further research. 
\textbf{Note:} The descriptions of all categories in the training set can be obtained through one single pre-processing step. Therefore, in \textbf{c. \& d.}, the additional computational cost during training can be ignored. 
More details of the templates and the designs for GPT and Llama-2 are provided in the \textit{Appendix}.

\begin{table}
\caption{\textbf{Ablations of the Representation Compensation strategy.} The Content-Dependent Transfer is removed. The best results are highlighted with \textcolor{red}{red}, and the default settings are highlighted with \textcolor{gray}{gray} background.}
\label{tab:RC}
  \begin{subtable}{0.4\textwidth}
    \centering

      \begin{tabular}{l|lllll}
      \Xhline{0.7pt}
      & \textbf{A-847} & \textbf{A-150} & \textbf{PC-59}\\ 
      \hline
      \rowcolor{gray!10}\multicolumn{1}{c|}{None} &  \textcolor{red}{14.6} & \textcolor{red}{34.8} & \textcolor{red}{57.1} \\     
      Freeze \{S0, 1\}&  \textcolor{red}{14.6} & 34.7 &  57.0 \\     
      Freeze \{S0, 1, 2\}& 14.0 & 34.6 & 55.3 \\     
      Freeze \{S0, 1, 2, 3\}& 13.6 & 33.6 & 54.7   \\     
      \Xhline{0.7pt}
      \end{tabular}

    \caption{Ablation of the \textbf{frozen stages} in CLIP-V.}
    \label{tab:RC-unit}
  \end{subtable}%
    \hspace{18mm}
  \begin{subtable}{0.4\textwidth}
    \centering

      \begin{tabular}{l|lllll}
      \Xhline{0.7pt}
      & \textbf{A-847} & \textbf{A-150} & \textbf{PC-59}\\ 
      \hline
      Grid \{1\}&  13.8 & 33.9 &  56.5 \\     
      Grid \{1, 2\}& 14.0 & 34.6 & 56.6 \\     
      \rowcolor{gray!10}\multicolumn{1}{c|}{Grid \{1, 2, 4\}}& \textcolor{red}{14.6} & \textcolor{red}{34.8} & \textcolor{red}{57.1} \\     
      Grid \{1, 3, 6\}& 14.5 & 34.6 & 55.7 \\     
      \Xhline{0.7pt}
      \end{tabular}

    \caption{Ablation of the $\mathrm{AvgPooling}$ \textbf{grid} in $\mathcal{L}_{rc}$.}
    \label{tab:RC-grid}
  \end{subtable}
\end{table}

\noindent \textbf{Effect of Representation Compensation.} We conduct ablation studies on Representation Compensation strategy in Fig. \ref{tab:RC}, here we remove the Content-Dependent Transfer for analysis.

\begin{itemize}[itemsep=2pt,topsep=0pt,parsep=0pt]
\item \textbf{Frozen stages in CLIP-V:}  We explore the impact of fine-tuning units within CLIP-V. CLIP-V consists of 4 ConvNeXt stages \{S0, S1, S2, S3\}, which downsample the image features from $\frac{1}{4}$ to $\frac{1}{32}$. 
We start with fine-tuning the entire CLIP-V, and then freezing each stage sequentially, as detailed in Tab. \ref{tab:RC-unit}. 
Compared to fine-tuning the entire CLIP-V, freezing any stage causes performance degradation. Freezing {S0-1}, {S0-2}, {S0-3} brings -0.1, -1.8, and -2.4 mIoU performance degradation respectively on PC-59, indicating that freezing S2 and S3 (depth convnext stages) has the most significant impact on the performance.
\item \textbf{Effect of $\mathrm{AvgPooling}$ grids:} In Tab. \ref{tab:RC-grid}, we investigate how different multi-scale $\mathrm{AvgPooling}$ grids ($\{1\}$, $\{1,2\}$, $\{1,2,4\}$, $\{1,3,6\}$) in $\mathcal{L}_{rc}$ impact performance. Results show $\{1,2,4\}$ grids boost performance on A-150 to 34.8 mIoU, and achieve the best performance. Using $\{1,3,6\}$ grads results in -1.6 drops on PC-59, manifesting overly large $\mathrm{AvgPooling}$ grids compromises the model to learn region-level differences. 
\end{itemize}

\begin{table}
  \centering
  \footnotesize
  \caption{Extending MAFT+ with ConvNeXt-Base CLIP. The best results are highlighted with \textcolor{red}{red}.}
% \resizebox{1.0\textwidth}{!}{
  \vspace{-3mm}

    \renewcommand\arraystretch{1.05} % 1.95
    
    \begin{tabular}{l|ccccc}
      \Xhline{0.7pt}

      & \textbf{~A-847~} & \textbf{~A-150~} & \textbf{~PC-459~} & \textbf{~PC-59~} & \textbf{~PAS-20~}\\ 
      \hline
       % MAFT \cite{maft} & \multirow{3}{*}{R50} & 8.4 & 27.0 & 9.9 & 50.8 & 89.0 \\
       FC-CLIP*\cite{fcclip} &  12.7  & 31.1 & 12.5  & 54.3 & 93.8 \\
      MAFT +  & \textcolor{red}{13.2}  & \textcolor{red}{33.6} & \textcolor{red}{14.2}  & \textcolor{red}{55.9} & \textcolor{red}{93.9} \\
      \Xhline{0.7pt}
      \end{tabular}
      % }
      \label{tab:vlms}

 \end{table}

\noindent \textbf{Extending MAFT+ with ConvNeXt-Base CLIP.} 
To showcase the efficacy and robustness of MAFT+, we conduct experiments using ConvNeXt-Base CLIP. The results are shown in Tab. \ref{tab:vlms}, we also include the results of FC-CLIP for comparison. 
Compared with the FC-CLIP counterpart, MAFT+ outperforms it by a significant margin on all five datasets. This demonstrates that MAFT+ can easily transfer to other CLIP models.

% \subsection{Extending MAFT+ with more Vision-Language Models}
% \input{tables/vlms}

% \noindent \textbf{CLIP-ViT-L.}
% According to Tab. \ref{tab:vlms}, FreeSeg with a standard CLIP-ViT-L model (dubbed $\mathrm{FreeSeg}$) still can not achieve satisfactory results. However, by integrating our MAFT (dubbed $\mathrm{FreeSeg+MAFT}$), the segmentation results are remarkably enhanced, thus establishing new state-of-the-art benchmarks.

% \noindent \textbf{CLIP-Res50.}
% Our MAFT can easily adapted into ResNet-based models. Specifically, we modified the $\mathrm{AttentionPool2d}$ unit within CLIP-R50 Image Encoder. The mask proposals are introduced as attention bias ($B$) in Multihead Attention, with $F_{cls}$ being repeated N times. Notably in CLIP-R50, $F_{cls}$ is obtained via $\mathrm{GlobalAveragePooling}$ performing on $F_{feat}$. The results are presented in Tab. \ref{tab:vlms}. The performance on all 5 datasets is improved by a large margin. $\mathrm{FreeSeg+MAFT}$ with CLIP-R50 achieves competitive results with some CLIP-ViT-B-based methods according to Tab. \ref{tab:ovs}.

\subsection{Qualitative Study}

\noindent \textbf{Visualizations of similarity map.}
\begin{figure}[t]
% \vspace{-10mm}
\begin{center}
   \includegraphics[width=0.9\linewidth]{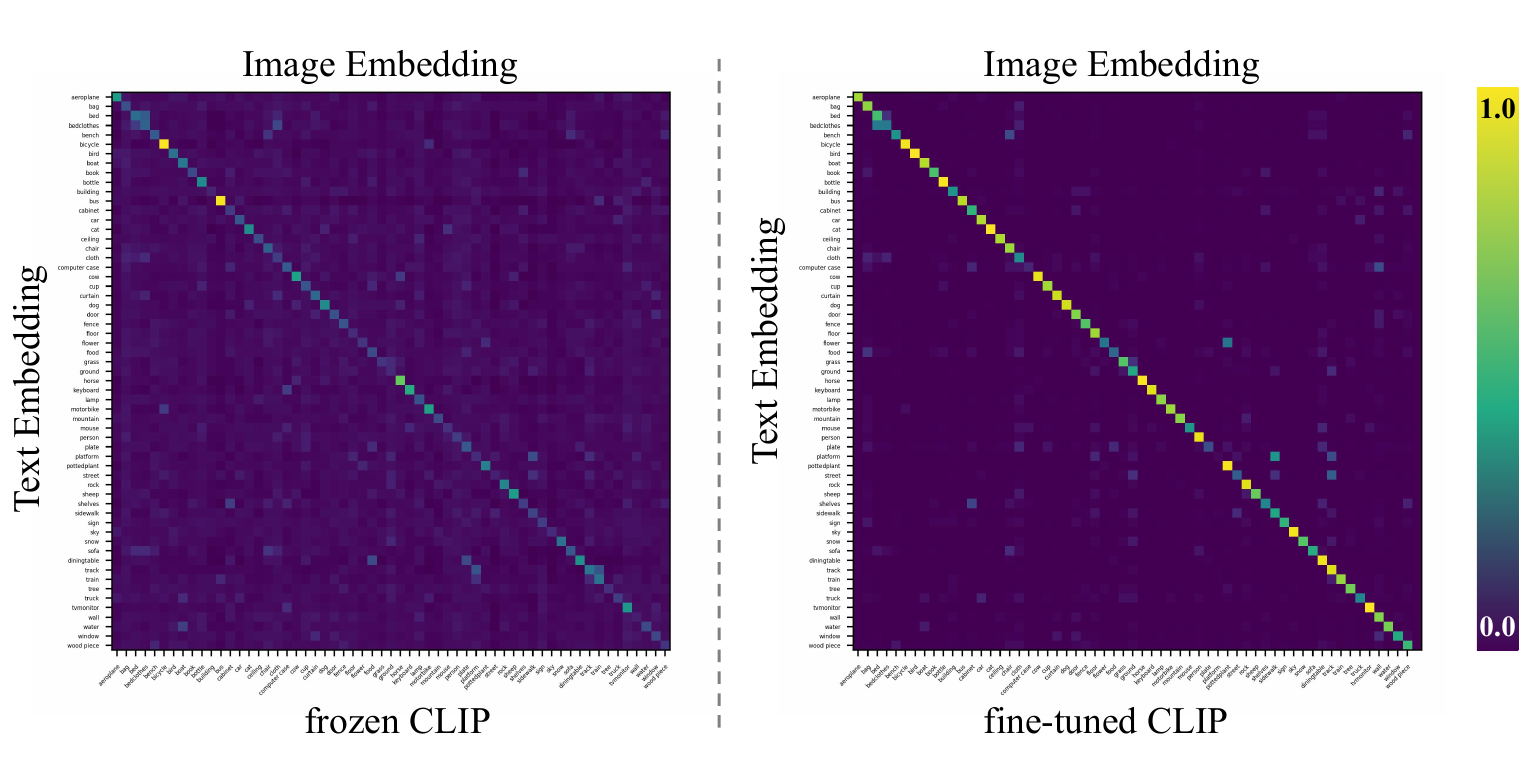}
\end{center}
\vspace{-6mm}
   \caption{
   Qualitative results. Normalized cosine similarity between the text embeddings and image embeddings of 59 classes in PC59. Text \& image embeddings are generated by frozen CLIP (left). Text \& image embeddings are generated by our MAFT+ fine-tuned CLIP (right). The high similarity scores are highlighted in yellow, low similarity scores are shown in blue. 
   }
\label{fig:vis-distance}
% \vspace{-5mm}
\end{figure}

Fig. \ref{fig:vis-distance} presents the normalized similarity map between text and image embeddings.
Including similarity map generated by frozen CLIP embeddings (left) and similarity map generated by fine-tuned CLIP embeddings (right). 
An observation can be obtained: The high similarity values of fine-tuned CLIP are mainly located on the diagonal of the similarity map, indicating the collaborative optimization of CLIP-V and CLIP-T achieves better alignment of vision-text representation.

\noindent \textbf{Qualitative analysis.}
\begin{figure}[t]
% \vspace{-10mm}
\begin{center}
   \includegraphics[width=0.95\linewidth]{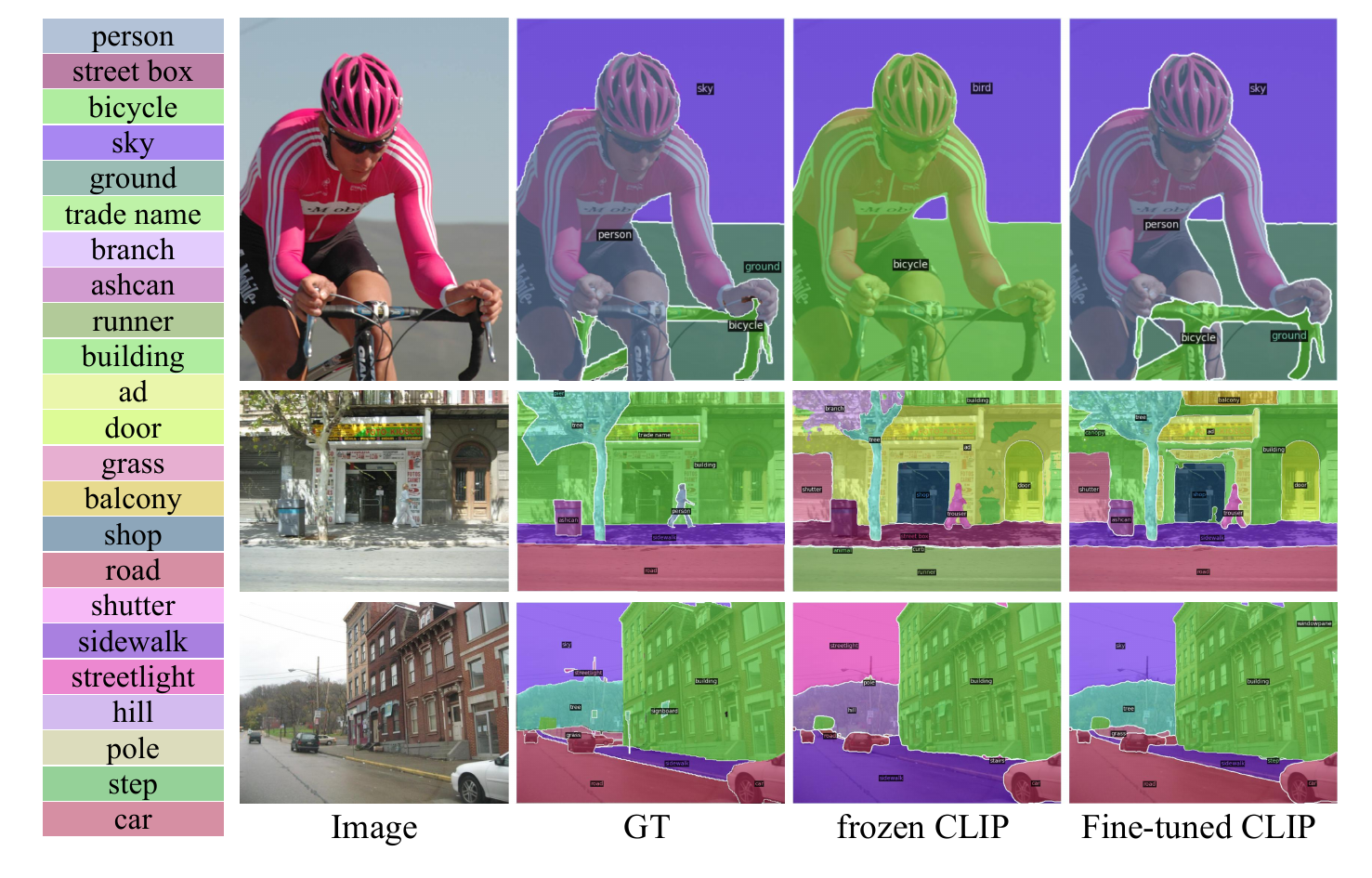}
\end{center}
\vspace{-5mm}
   \caption{
   Qualitative results. The results with the frozen CLIP and our MAFT+ fine-tuned CLIP are shown for comparasion.
   }
\label{fig:vis-final}
% \vspace{-5mm}
\end{figure}

We show some visual examples in Fig. \ref{fig:vis-final}. In some simple cases, the frozen CLIP results may contain background noise, and tend to classify multiple objects into one single class (\textit{e.g.} the $1^{st}$ row, ``bicycle''). The frozen CLIP is prone to misclassification when there are many categories in one image (the $3^{rd}$ row, ``streetlight'', ``sidewalk'', ``hill''). 
Our fine-tuned CLIP collaboratively learns vision-text representation for segmentation tasks, which can significantly improve the segmentation results.
In addition, the $2^{rd}$ row shows that our fine-tuned CLIP successfully segments ``balcony'', which is a reasonable outcome even though ``balcony'' does not appear in the ground-truth annotations. 
% \textcolor{blue}{This showcases the model's ability to capture meaningful semantic information beyond the provided annotations.}
% ``Please describe the appearance of $cat$.''
More visual samples are shown in the \textit{Appendix}.

\section{Conclusion}
In this paper, we rethink the issues in frozen CLIP paradigm and CLIP-V fine-tuning paradigm and propose a collaborative vision-text optimizing structure, MAFT+, for OVS.
We introduce the Representation Compensation strategy to review the original CLIP's representation to maintain the zero-shot capability of CLIP-V. And propose the Content-Dependent Transfer to optimize the text representation in a parameter-efficient way. Extensive experiments well demonstrate our MAFT+ achieves superior performance on multiple open-vocabulary segmentation datasets.

\textbf{Limitations.}
% Although our MAFT+ optimizes the fine-tuning process of CLIP and avoids the loss of CLIP generalization through representation compensation. But in fact, with the fine-tuning of MAFT+, the generalization of CLIP will still decline.
% How to maintain the zero-shot capability of CLIP during the fine-tune process is our future research focus.
While the proposed MAFT+ optimizes the vision-test representation space of CLIP to fit the distribution of OVS, it is important to acknowledge that the optimization upper-bound is constrained by the capabilities of the pre-trained CLIP model. Addressing this limitation constitute our future research focus.

\section*{Acknowledgements}
This work was supported in part by the National Key R \& D Program of China (No. 2021ZD0112100), the National NSF of China (No.U23A20314).

\newpage

\bibliographystyle{splncs04}
\bibliography{main}

\begin{thebibliography}{10}
\providecommand{\url}[1]{\texttt{#1}}
\providecommand{\urlprefix}{URL }
\providecommand{\doi}[1]{https://doi.org/#1}

\bibitem{gpt}
Brown, T., Mann, B., Ryder, N., Subbiah, M., Kaplan, J.D., Dhariwal, P., Neelakantan, A., Shyam, P., Sastry, G., Askell, A., et~al.: Language models are few-shot learners. Advances in neural information processing systems  \textbf{33},  1877--1901 (2020)

\bibitem{zs5}
Bucher, M., Vu, T.H., Cord, M., P{\'e}rez, P.: Zero-shot semantic segmentation. Advances in Neural Information Processing Systems  \textbf{32} (2019)

\bibitem{coco}
Caesar, H., Uijlings, J., Ferrari, V.: Coco-stuff: Thing and stuff classes in context. In: Proceedings of the IEEE/CVF Conference on Computer Vision and Pattern Recognition. pp. 1209--1218 (2018)

\bibitem{chen2017deeplab}
Chen, L.C., Papandreou, G., Kokkinos, I., Murphy, K., Yuille, A.L.: Deeplab: Semantic image segmentation with deep convolutional nets, atrous convolution, and fully connected crfs. IEEE transactions on pattern analysis and machine intelligence  \textbf{40}(4),  834--848 (2017)

\bibitem{opsnet}
Chen, X., Li, S., Lim, S.N., Torralba, A., Zhao, H.: Open-vocabulary panoptic segmentation with embedding modulation. arXiv preprint arXiv:2303.11324  (2023)

\bibitem{cheng2021mask2former}
Cheng, B., Choudhuri, A., Misra, I., Kirillov, A., Girdhar, R., Schwing, A.G.: Mask2former for video instance segmentation. arXiv preprint arXiv:2112.10764  (2021)

\bibitem{cheng2021maskformer}
Cheng, B., Schwing, A., Kirillov, A.: Per-pixel classification is not all you need for semantic segmentation. Advances in Neural Information Processing Systems  \textbf{34} (2021)

\bibitem{zegformer}
Ding, J., Xue, N., Xia, G.S., Dai, D.: Decoupling zero-shot semantic segmentation. In: Proceedings of the IEEE/CVF Conference on Computer Vision and Pattern Recognition. pp. 11583--11592 (2022)

\bibitem{pascal}
Everingham, M., Eslami, S.A., Van~Gool, L., Williams, C.K., Winn, J., Zisserman, A.: The pascal visual object classes challenge: A retrospective. International journal of computer vision  \textbf{111},  98--136 (2015)

\bibitem{fang2023locating}
Fang, Y., Zhu, F., Cheng, B., Liu, L., Zhao, Y., Wei, Y.: Locating noise is halfway denoising for semi-supervised segmentation. In: Proceedings of the IEEE/CVF International Conference on Computer Vision. pp. 16612--16622 (2023)

\bibitem{ghiasi2022scaling}
Ghiasi, G., Gu, X., Cui, Y., Lin, T.Y.: Scaling open-vocabulary image segmentation with image-level labels. In: Computer Vision--ECCV 2022: 17th European Conference, Tel Aviv, Israel, October 23--27, 2022, Proceedings, Part XXXVI. pp. 540--557. Springer (2022)

\bibitem{cagnet}
Gu, Z., Zhou, S., Niu, L., Zhao, Z., Zhang, L.: Context-aware feature generation for zero-shot semantic segmentation. In: Proceedings of the 28th ACM International Conference on Multimedia. pp. 1921--1929 (2020)

\bibitem{han2023global}
Han, K., Liu, Y., Liew, J.H., Ding, H., Wei, Y., Liu, J., Wang, Y., Tang, Y., Yang, Y., Feng, J., et~al.: Global knowledge calibration for fast open-vocabulary segmentation. In: Proceedings of the IEEE/CVF International Conference on Computer Vision (2023)

\bibitem{huang2019ccnet}
Huang, Z., Wang, X., Huang, L., Huang, C., Wei, Y., Liu, W.: Ccnet: Criss-cross attention for semantic segmentation. In: Proceedings of the IEEE/CVF International Conference on Computer Vision. pp. 603--612 (2019)

\bibitem{huang2021alignseg}
Huang, Z., Wei, Y., Wang, X., Liu, W., Huang, T.S., Shi, H.: Alignseg: Feature-aligned segmentation networks. IEEE Transactions on Pattern Analysis and Machine Intelligence  \textbf{44}(1),  550--557 (2021)

\bibitem{ilharco2openclip}
Ilharco, G., Wortsman, M., Wightman, R., Gordon, C., Carlini, N., Taori, R., Dave, A., Shankar, V., Namkoong, H., Miller, J., et~al.: Openclip, july 2021. If you use this software, please cite it as below  \textbf{2}(4), ~5 (2021)

\bibitem{jia2021scaling}
Jia, C., Yang, Y., Xia, Y., Chen, Y.T., Parekh, Z., Pham, H., Le, Q., Sung, Y.H., Li, Z., Duerig, T.: Scaling up visual and vision-language representation learning with noisy text supervision. In: International Conference on Machine Learning. pp. 4904--4916. PMLR (2021)

\bibitem{maft}
Jiao, S., Wei, Y., Wang, Y., Zhao, Y., Shi, H.: Learning mask-aware clip representations for zero-shot segmentation. Advances in Neural Information Processing Systems  \textbf{36} (2023)

\bibitem{kirillov2019panoptic}
Kirillov, A., He, K., Girshick, R., Rother, C., Doll{\'a}r, P.: Panoptic segmentation. In: Proceedings of the IEEE/CVF conference on computer vision and pattern recognition. pp. 9404--9413 (2019)

\bibitem{kuhn1955hungarian}
Kuhn, H.W.: The hungarian method for the assignment problem. Naval research logistics quarterly  \textbf{2}(1-2),  83--97 (1955)

\bibitem{li2022languagedriven}
Li, B., Weinberger, K.Q., Belongie, S., Koltun, V., Ranftl, R.: Language-driven semantic segmentation. In: International Conference on Learning Representations (2022), \url{https://openreview.net/forum?id=RriDjddCLN}

\bibitem{ovseg}
Liang, F., Wu, B., Dai, X., Li, K., Zhao, Y., Zhang, H., Zhang, P., Vajda, P., Marculescu, D.: Open-vocabulary semantic segmentation with mask-adapted clip. In: Proceedings of the IEEE/CVF Conference on Computer Vision and Pattern Recognition. pp. 7061--7070 (2023)

\bibitem{lin2014microsoft}
Lin, T.Y., Maire, M., Belongie, S., Hays, J., Perona, P., Ramanan, D., Doll{\'a}r, P., Zitnick, C.L.: Microsoft coco: Common objects in context. In: Computer Vision--ECCV 2014: 13th European Conference, Zurich, Switzerland, September 6-12, 2014, Proceedings, Part V 13. pp. 740--755. Springer (2014)

\bibitem{mccloskey1989catastrophic}
McCloskey, M., Cohen, N.J.: Catastrophic interference in connectionist networks: The sequential learning problem. In: Psychology of learning and motivation, vol.~24, pp. 109--165. Elsevier (1989)

\bibitem{pc}
Mottaghi, R., Chen, X., Liu, X., Cho, N.G., Lee, S.W., Fidler, S., Urtasun, R., Yuille, A.: The role of context for object detection and semantic segmentation in the wild. In: Proceedings of the IEEE/CVF Conference on Computer Vision and Pattern Recognition. pp. 891--898 (2014)

\bibitem{freeseg}
Qin, J., Wu, J., Yan, P., Li, M., Yuxi, R., Xiao, X., Wang, Y., Wang, R., Wen, S., Pan, X., et~al.: Freeseg: Unified, universal and open-vocabulary image segmentation. arXiv preprint arXiv:2303.17225  (2023)

\bibitem{radford2021learning}
Radford, A., Kim, J.W., Hallacy, C., Ramesh, A., Goh, G., Agarwal, S., Sastry, G., Askell, A., Mishkin, P., Clark, J., et~al.: Learning transferable visual models from natural language supervision. In: International conference on machine learning. pp. 8748--8763. PMLR (2021)

\bibitem{rombach2022high}
Rombach, R., Blattmann, A., Lorenz, D., Esser, P., Ommer, B.: High-resolution image synthesis with latent diffusion models. In: Proceedings of the IEEE/CVF conference on computer vision and pattern recognition. pp. 10684--10695 (2022)

\bibitem{shaban2017one}
Shaban, A., Bansal, S., Liu, Z., Essa, I., Boots, B.: One-shot learning for semantic segmentation. arXiv preprint arXiv:1709.03410  (2017)

\bibitem{svf}
Sun, Y., Chen, Q., He, X., Wang, J., Feng, H., Han, J., Ding, E., Cheng, J., Li, Z., Wang, J.: Singular value fine-tuning: Few-shot segmentation requires few-parameters fine-tuning. arXiv preprint arXiv:2206.06122  (2022)

\bibitem{touvron2023llama}
Touvron, H., Lavril, T., Izacard, G., Martinet, X., Lachaux, M.A., Lacroix, T., Rozi{\`e}re, B., Goyal, N., Hambro, E., Azhar, F., et~al.: Llama: Open and efficient foundation language models. arXiv preprint arXiv:2302.13971  (2023)

\bibitem{spnet}
Xian, Y., Choudhury, S., He, Y., Schiele, B., Akata, Z.: Semantic projection network for zero-and few-label semantic segmentation. In: Proceedings of the IEEE/CVF Conference on Computer Vision and Pattern Recognition. pp. 8256--8265 (2019)

\bibitem{conti-ema}
Xiao, J.W., Zhang, C.B., Feng, J., Liu, X., van~de Weijer, J., Cheng, M.M.: Endpoints weight fusion for class incremental semantic segmentation. In: Proceedings of the IEEE/CVF Conference on Computer Vision and Pattern Recognition. pp. 7204--7213 (2023)

\bibitem{xu2022groupvit}
Xu, J., De~Mello, S., Liu, S., Byeon, W., Breuel, T., Kautz, J., Wang, X.: Groupvit: Semantic segmentation emerges from text supervision. CVPR  (2022)

\bibitem{odise}
Xu, J., Liu, S., Vahdat, A., Byeon, W., Wang, X., De~Mello, S.: Open-vocabulary panoptic segmentation with text-to-image diffusion models. In: Proceedings of the IEEE/CVF Conference on Computer Vision and Pattern Recognition. pp. 2955--2966 (2023)

\bibitem{xu2024transferable}
Xu, J., Chen, W., Zhao, Y., Wei, Y.: Transferable and principled efficiency for open-vocabulary segmentation. In: Proceedings of the IEEE/CVF Conference on Computer Vision and Pattern Recognition. pp. 15814--15824 (2024)

\bibitem{san}
Xu, M., Zhang, Z., Wei, F., Hu, H., Bai, X.: Side adapter network for open-vocabulary semantic segmentation. In: Proceedings of the IEEE/CVF Conference on Computer Vision and Pattern Recognition. pp. 2945--2954 (2023)

\bibitem{zsseg}
Xu, M., Zhang, Z., Wei, F., Lin, Y., Cao, Y., Hu, H., Bai, X.: A simple baseline for open-vocabulary semantic segmentation with pre-trained vision-language model. In: Computer Vision--ECCV 2022: 17th European Conference, Tel Aviv, Israel, October 23--27, 2022, Proceedings, Part XXIX. pp. 736--753. Springer (2022)

\bibitem{fcclip}
Yu, Q., He, J., Deng, X., Shen, X., Chen, L.C.: Convolutions die hard: Open-vocabulary segmentation with single frozen convolutional clip. Advances in Neural Information Processing Systems  \textbf{36} (2023)

\bibitem{conti-dis}
Zhang, C.B., Xiao, J.W., Liu, X., Chen, Y.C., Cheng, M.M.: Representation compensation networks for continual semantic segmentation. In: Proceedings of the IEEE/CVF Conference on Computer Vision and Pattern Recognition. pp. 7053--7064 (2022)

\bibitem{mmformer}
Zhang, G., Navasardyan, S., Chen, L., Zhao, Y., Wei, Y., Shi, H., et~al.: Mask matching transformer for few-shot segmentation. Advances in Neural Information Processing Systems  \textbf{35},  823--836 (2022)

\bibitem{zhang2023slca}
Zhang, G., Wang, L., Kang, G., Chen, L., Wei, Y.: Slca: Slow learner with classifier alignment for continual learning on a pre-trained model. In: Proceedings of the IEEE/CVF International Conference on Computer Vision. pp. 19148--19158 (2023)

\bibitem{zhang2023controlvideo}
Zhang, Y., Wei, Y., Jiang, D., Zhang, X., Zuo, W., Tian, Q.: Controlvideo: Training-free controllable text-to-video generation. arXiv preprint arXiv:2305.13077  (2023)

\bibitem{zhang2022mining}
Zhang, Z., Gao, G., Fang, Z., Jiao, J., Wei, Y.: Mining unseen classes via regional objectness: A simple baseline for incremental segmentation. Advances in Neural Information Processing Systems  \textbf{35},  24340--24353 (2022)

\bibitem{zhangcoinseg}
Zhang, Z., Gao, G., Jiao, J., Liu, C.H., Wei, Y.: Coinseg: Contrast inter-and intra-class representations for incremental segmentation. In: Proceedings of the IEEE/CVF International Conference on Computer Vision (2023)

\bibitem{pspnet}
Zhao, H., Shi, J., Qi, X., Wang, X., Jia, J.: Pyramid scene parsing network. In: Proceedings of the IEEE/CVF Conference on Computer Vision and Pattern Recognition. pp. 2881--2890 (2017)

\bibitem{ding2023maskclip}
Zheng~Ding, Jieke~Wang, Z.T.: Open-vocabulary universal image segmentation with maskclip. In: International Conference on Machine Learning (2023)

\bibitem{ade20k}
Zhou, B., Zhao, H., Puig, X., Fidler, S., Barriuso, A., Torralba, A.: Scene parsing through ade20k dataset. In: Proceedings of the IEEE/CVF Conference on Computer Vision and Pattern Recognition. pp. 633--641 (2017)

\end{thebibliography}

\newpage

\appendix
\section*{Appendix}

We first introduce the dataset settings in Sec. \ref{sec:dataset}. Then, the Prompt engineering techenic is introduced in detailed in Sec. \ref{sec:tempt}, including the template-based Prompt and the description-based Prompt. 
Moreover, we provide additional qualitative results in Sec. \ref{sec:vis}.

% lin2014microsoft
\section{Dataset}
\label{sec:dataset}   
We follow \cite{san, fcclip, odise, maft} to conduct experiments on the popular benchmarks of open-vocabulary \textit{semantic} and \textit{panoptic} settings, COCO-Stuff \cite{coco}, COCO-Panoptic \cite{lin2014microsoft}, Pascal-VOC \cite{pascal} ADE20K \cite{ade20k}, and Pascal-Context \cite{pc} to evaluate the performance of MAFT+. 

\begin{itemize}[itemsep=2pt,topsep=0pt,parsep=0pt]
\item \textbf{COCO-Stuff}: COCO-Stuff is a large-scale semantic segmentation dataset that contains 164K images with 171 annotated classes, which are divided into the training set (118K images), validation set (5K images), and testing set (41K images). In our experiments, we use the full 118K training set as the training data to train the \textit{semantic} models.

\item \textbf{COCO-Panoptic}: COCO-Panoptic shares the same training images with COCO-Stuff. These images are labeled into 133 categories. In our experiments, we use COCO-Panoptic to train the \textit{panoptic} models.

\item \textbf{Pascal-VOC}: Pascal-VOC includes 1,449 images for testing with 20 annotated classes. In the open-vocabulary \textit{semantic} segmentation, all 20 classes are used for evaluation (dubbed as PAS-20).

\item \textbf{ADE20K}: ADE20K is a large-scale scene understanding dataset comprising 2k images for validation with two types of annotations: one with 150 classes featuring panoptic annotations and another with 847 classes featuring semantic annotations.  For the open-vocabulary \textit{semantic} segmentation, we evaluate our method on two settings of ADE20K: 150 classes (dubbed as A-150) and 847 classes (dubbed as A-847). In the open-vocabulary \textit{panoptic} segmentation, we use the setting with 150 class annotations for evaluation.

\item \textbf{Pascal-Context} is a dataset for semantic understanding which contains 5K validation images. Two versions are used for open-vocabulary \textit{semantic} segmentation, one with 59 frequently used classes (dubbed as PC-59) and another with the whole 459 classes (dubbed as PC-459).

\end{itemize}

\section{Template-based Prompt \& Description-based Prompt}
\label{sec:tempt}

Prompt engineering has been proven to be beneficial for open-vocabulary segmentation.  In our default setting, we follow the common practice of using the template-based prompt to augment class names into sentences. In addition, we also explore using GPT \cite{gpt} or Llama \cite{touvron2023llama} to apply description-based prompts.

\noindent \textbf{Template-based Prompt.}
Following established approaches \cite{san, fcclip, odise, maft}, we use multiple templates to integrate the class names into sentences. These sentences are then fed into CLIP-T, and the resulting outputs are averaged to generate the text embedding for each class. The templates are listed in Tab. \ref{tab:template}.

\noindent \textbf{Description-based Prompt.}
We assume that the detailed descriptions of one class name contain additional valuable information that helps to optimize CLIP-T. To investigate this, we design description-based prompts, leveraging Large Language Models (LLMs) to generate descriptions, including using GPT-3.5 \cite{gpt} to generate description sentences, and use the open-source LLM, Llama-2 \cite{touvron2023llama}, to generate descriptive text embeddings. Through experimental verification, we selected a few prompts suitable for LLMs to generate descriptions. The prompts and responses are shown in Tab. \ref{tab:descript1} and Tab. \ref{tab:descript2}, respectively.

The results indicate that some descriptions provide valuable visual attributes, facilitating the alignment of vision-text representations in the CLIP feature space. However,  they may introduce noise. \textit{e.g.}, both \textit{cat} and \textit{chair} have descriptions that include the sentence “four legs”.

\section{Similarity Map \& Visualize results}
\label{sec:vis}

\begin{figure}
% \vspace{-10mm}
\begin{center}
   \includegraphics[width=0.95\linewidth]{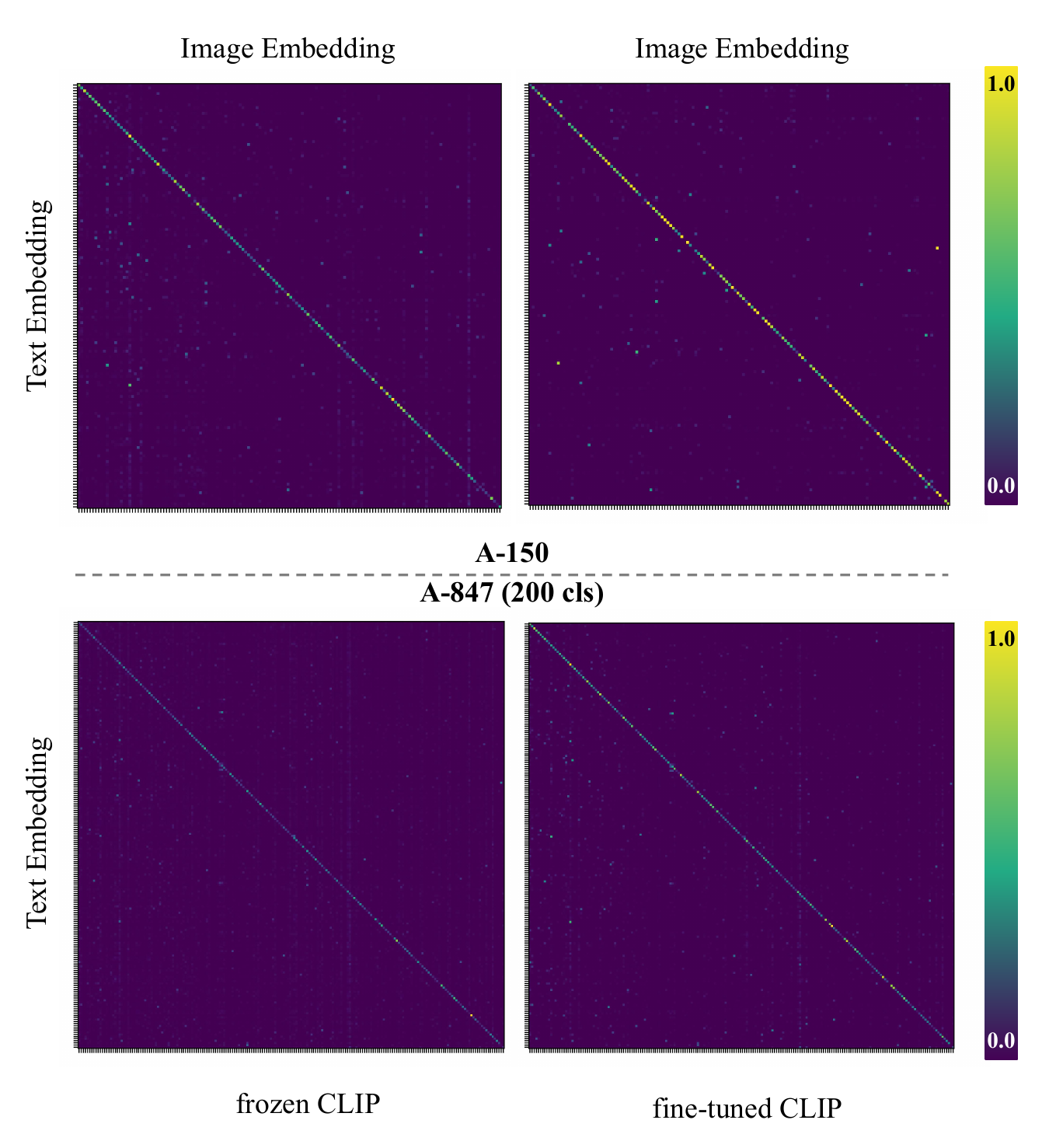}
\end{center}
\vspace{-5mm}
   \caption{
   Normalized cosine similarity on A-150 and A-847, we choose 200 categories in A-847 for visualization. }
\label{fig:appendix-similar}
% \vspace{-5mm}
\end{figure}

\begin{figure}
% \vspace{-10mm}
\begin{center}
   \includegraphics[width=0.95\linewidth]{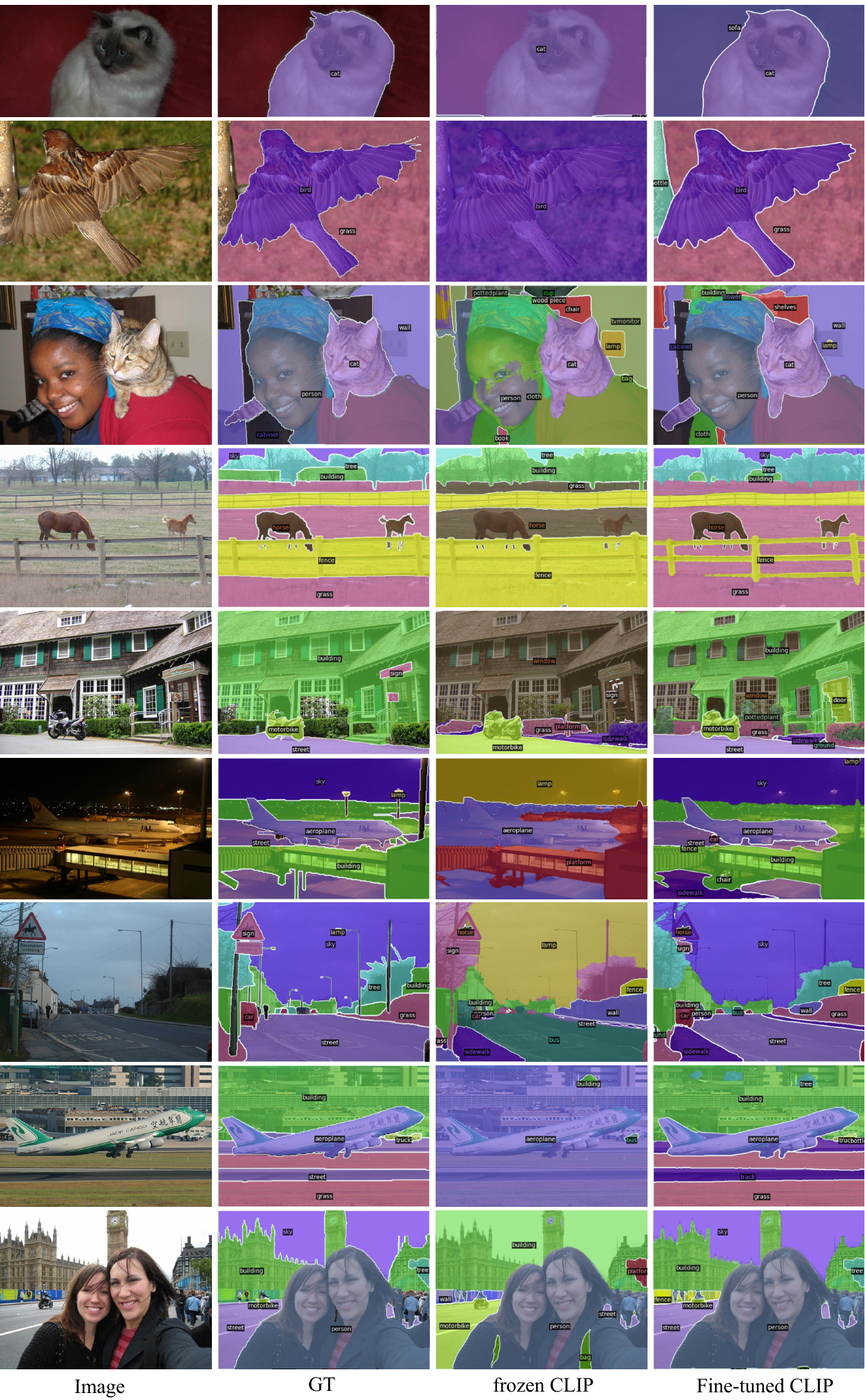}
\end{center}
\vspace{-5mm}
   \caption{
   Qualitative results. }
\label{fig:appendix-vis1}
% \vspace{-5mm}
\end{figure}

\begin{figure}
% \vspace{-10mm}
\begin{center}
   \includegraphics[width=0.95\linewidth]{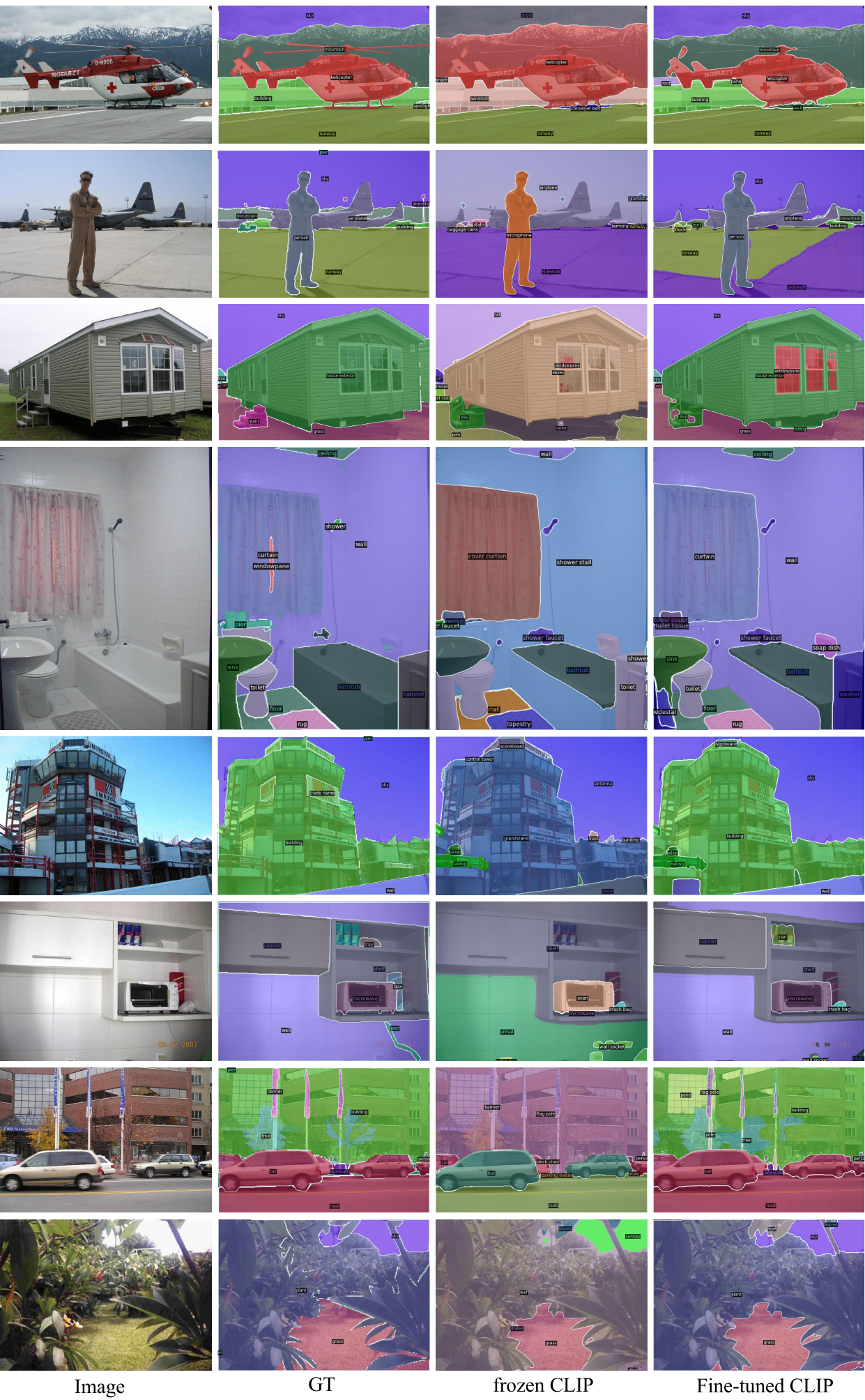}
\end{center}
\vspace{-5mm}
   \caption{
   Qualitative results. }
\label{fig:appendix-vis2}
% \vspace{-5mm}
\end{figure}

We provide more qualitative results, including similarity maps (Fig. \ref{fig:appendix-similar}), and visualize results in Pascal-VOC, COCO-Stuff, ADE20K datasets (Fig. \ref{fig:appendix-vis1}, \ref{fig:appendix-vis2}).

\noindent \textbf{Similarity map.}
Fig. \ref{fig:appendix-similar} presents the normalized similarity maps between text and image embeddings in A-150 and A-847 datasets. We choose 200 categories in A-847 for visualization. It is evident that the elevated similarity values of fine-tuned CLIP's similarity map are mainly located on the main diagonal, indicating the fine-tuned CLIP achieves a better alignment of vision-text representation.

\noindent \textbf{Qualitative Analysis.} Fig. \ref{fig:appendix-vis1}, \ref{fig:appendix-vis2} show segmentation results on Pascal-VOC, COCO-Stuff, ADE20K. The frozen CLIP results may contain background noise (1st and 2nd rows in Fig. \ref{fig:appendix-vis1}) or misclassify when there are many objects in one image (3rd row in Fig. \ref{fig:appendix-vis1}). The fine-tuned CLIP generates better results compared to the frozen CLIP, which can even correct misclassified areas in ground-truth (\textit{fence} in the 4th row in Fig. \ref{fig:appendix-vis1}).

\begin{table}
  \centering
  \scriptsize  % scriptsize   footnotesize
  % \vspace{-10pt}
  \caption{Prompt templates used in our method.}
    \renewcommand\arraystretch{1.1} % 1.95
    \begin{tabular}{c}
      % \toprule
      \Xhline{0.7pt}

     Templates\\ 
      \hline
    “a photo of a $\{~\}$.” \\
    “This is a photo of a $\{~\}$” \\
    “There is a $\{~\}$ in the scene” \\
    “There is the $\{~\}$ in the scene” \\
    “a photo of a $\{~\}$ in the scene” \\
    “a photo of a small $\{~\}$.” \\
    “a photo of a medium $\{~\}$.” \\
    “a photo of a large $\{~\}$.” \\
    “This is a photo of a small $\{~\}$.” \\
    “This is a photo of a medium $\{~\}$.” \\
    “This is a photo of a large $\{~\}$.” \\
    “There is a small $\{~\}$ in the scene.” \\
    “There is a medium $\{~\}$ in the scene.” \\
    “There is a large $\{~\}$ in the scene.” \\

      \Xhline{0.7pt}
      % \bottomrule
      \end{tabular}
      % }
      % \end{threeparttable}
      \label{tab:template}
  % \vspace{-2mm}
  \end{table}

% \begin{table}
%   \centering
%   \scriptsize  % scriptsize   footnotesize
%   % \vspace{-10pt}
%   \caption{Description prompts used in our method.}
%     \renewcommand\arraystretch{1.0} % 1.95
%     \begin{tabular}{c}
%       % \toprule
%       \Xhline{0.7pt}

%      Description prompts\\ 
%       \hline
% "Please describe the appearance of $\{~\}$. Please characterize it briefly." \\
% "Describe the physical attributes of $\{~\}$. Please characterize it briefly." \\
% "What can you tell me about the appearance of the category of $\{~\}$? Please characterize it briefly." \\
% "Tell me about the outward features of the category of $\{~\}$. Please characterize it briefly." \\
% "Briefly outline the visual traits of the category of $\{~\}$." \\
% "Can you provide details about what the category of $\{~\}$ looks like? Please characterize it briefly." \\
% "I'm curious about the visual characteristics of the category of $\{~\}$. Please characterize it briefly." \\
% "Provide a description of the visual aspects of $\{~\}$. Please characterize it briefly." \\

%       \Xhline{0.7pt}
%       % \bottomrule
%       \end{tabular}
%       % }
%       % \end{threeparttable}
%       \label{tab:descript}
%   % \vspace{-2mm}
%   \end{table}  

\begin{table}
  \centering

  \caption{Description-based prompts and responses}
  \label{tab:descript}
  \begin{subtable}{1.0\textwidth}
    \centering

    \scriptsize  % scriptsize   footnotesize
    \renewcommand\arraystretch{1.1} % 1.95
    \begin{tabular}{c}
      % \toprule
      \Xhline{0.7pt}

     Description prompts\\ 
      \hline
“Please describe the appearance of $\{~\}$. Please characterize it briefly.” \\
“Describe the physical attributes of $\{~\}$. Please characterize it briefly.” \\
“What can you tell me about the appearance of the category of $\{~\}$? Please characterize it briefly.” \\
“Tell me about the outward features of the category of $\{~\}$. Please characterize it briefly.” \\
“Briefly outline the visual traits of the category of $\{~\}$.” \\
“Can you provide details about what the category of $\{~\}$ looks like? Please characterize it briefly.” \\
“I'm curious about the visual characteristics of the category of $\{~\}$. Please characterize it briefly.” \\
“Provide a description of the visual aspects of $\{~\}$. Please characterize it briefly.” \\
“Q: What are visual features of distinguishing a smartphone? A: - a touchscreen \\
Q: What are features for distinguishing a $\{~\}$?
A: -”\\

      \Xhline{0.7pt}
      % \bottomrule
      \end{tabular}
    \caption{Description prompts used in our method.}
    \label{tab:descript1}
  \end{subtable}

  \hfill
  \begin{subtable}{1.0\textwidth}
    \centering
    \includegraphics[width=\linewidth]{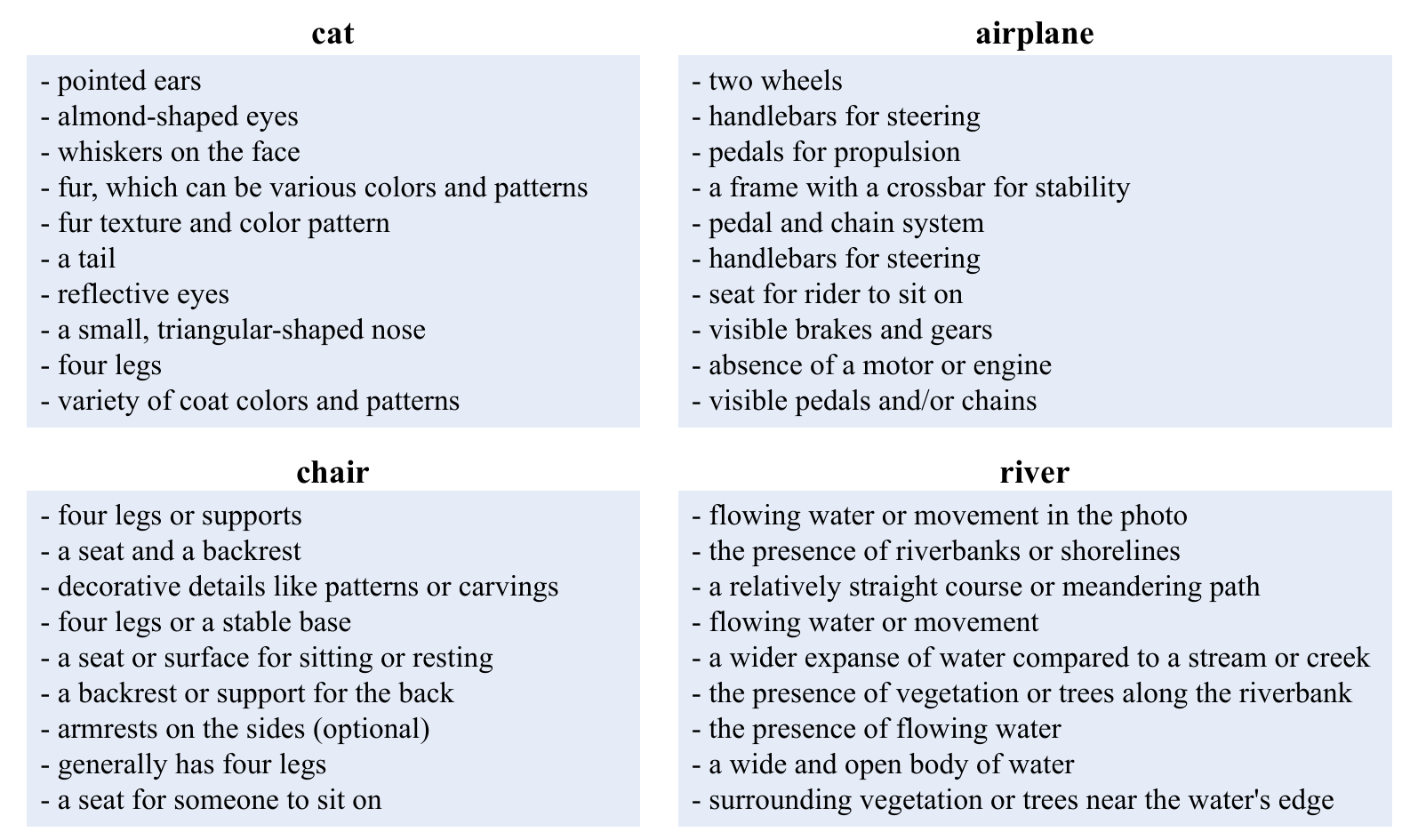}
    \caption{LLMs responses of category \textit{cat}, \textit{airplane}, \textit{chair}, and \textit{river}.}
    \label{tab:descript2}
  \end{subtable}

\end{table}

\clearpage

\end{document}